%% file: main.tex
\documentclass[10pt]{article}

\usepackage[margin=1in]{geometry}
\usepackage[T1]{fontenc}
\usepackage[utf8]{inputenc}
\usepackage{amsmath}
\usepackage{amssymb}
\usepackage{booktabs}
\usepackage{graphicx}
\usepackage{microtype}
\usepackage[round]{natbib}
\usepackage{placeins}
\usepackage{xcolor}
\usepackage{hyperref}

\hypersetup{
  colorlinks=true,
  linkcolor=blue!55!black,
  citecolor=blue!55!black,
  urlcolor=blue!55!black,
  pdftitle={Governance Records as Supervision: Verifier-Selected Self-Training for Structured Workflow Repair},
  pdfauthor={Jesus Salas (jesus.salas@gmail.com), Independent Researcher},
  pdfsubject={Verifier-admitted self-training, training-seed replication, and structured workflow repair}
}

\newcommand{\qwenmodel}{Qwen3-14B}
\newcommand{\planbench}{\textsc{PlanBench}}
\newcommand{\valsys}{\textsc{VAL}}

\title{Governance Records as Supervision:\\
Verifier-Selected Self-Training for Structured Workflow Repair}
\author{Jesus Salas\thanks{The author is employed by Microsoft. This work was
performed independently, outside the scope of that employment, on personal time
and equipment. It does not represent the views or positions of Microsoft
Corporation.}\\
Independent Researcher\\
\texttt{jesus.salas@gmail.com}}
\date{August 2026}

\begin{document}

\maketitle

\input{sections/00_abstract}
\input{sections/01_introduction}
\input{sections/02_problem}
\input{sections/03_method}
\input{sections/04_core_evidence}
\input{sections/08_discussion}
\input{sections/09_limitations}
\input{sections/10_related_work}
\input{sections/12_conclusion}

\bibliographystyle{plainnat}
\bibliography{references}

\appendix
\input{sections/appendix_planbench}
\input{sections/appendix_synthetic_program}

\end{document}

%% file: sections/00_abstract.tex
\begin{abstract}
Machine-verifiable workflows produce governance records linking a task
contract, model attempt, verifier decision, accepted output, and target origin.
We test whether verifier-admitted outputs can supervise a bounded model by
consolidating occasional or expensive capability into reliable one-shot
execution.

On fresh, structure-disjoint \planbench{} replanning cases, \qwenmodel{}
thinking produced 24 plans admitted by independently authored \valsys{}. They
trained the same checkpoint for non-thinking execution, without oracle targets
or a stronger teacher. \valsys{} acceptance rose from 1/80 to 57/80. A
prospective replication held targets, model revision, recipe, and evaluation
corpus fixed across eight LoRA seeds and three inference realizations per seed.
Every seed produced a clear lift: adapters reached 45/80 to 70/80 against 1/80
for every matched base report; the exact seed-level sign-flip test gave
$p=0.0078125$.

Target-selection performance was less stable. An initial matched seed gave
102/160 accepted plans after \valsys{} selection versus 69/160 after blinded
model self-selection. Across eight prospective seeds, the contrast was
seed-dependent, included one clear reverse seed, and did not replicate
($p=0.3672$). \valsys{} also had a positive descriptive aggregate over
schema-only selection but failed its preregistered seed-level reliability gate
($p=0.0703$). The verifier remains the admission authority; no reliable
downstream capability advantage of semantic selection is established.

A complementary Phi arm supports stronger-teacher distillation. Earlier
synthetic studies bound teachability, cumulative learning, transfer, and
stopping. The evidence supports robust consolidation of one fixed,
machine-checkable capability, not arbitrary planning, enterprise validity, or
unrestricted self-improvement.
\end{abstract}

%% file: sections/01_introduction.tex
\section{Introduction}

Language-model agents increasingly produce work whose validity is not purely
linguistic. A workflow may require every obligation exactly once, restrict the
tool assigned to each task, impose ordering and time windows, or satisfy an
executable planning model. The proposal is probabilistic, but many failures are
mechanically identifiable.

Most systems use this signal only at inference time: reject the invalid plan,
return an error, and try again. We ask whether the resulting runtime record can
also become a post-training asset. The basic transformation is:

\begin{equation}
  (\text{contract},\ \text{invalid plan},\ \text{verifier signal})
  \longrightarrow \text{verified target plan}.
\end{equation}

We call this tuple a \emph{governance record} when it preserves the declared
contract, model attempt, verifier result, accepted output, producer and student
lineage, and applicable gate outcome. If such records are reusable, low-rank
adaptation can consolidate recurring verified behavior into a bounded executor
while the external verifier remains the acceptance control.

This terminology builds on prior work that separates task correctness from
governed execution through decision, execution, and change provenance
\citep{salas2026correct}. That work treats provenance as auditable control
state. Here we isolate a narrower, machine-verifiable execution record and ask
whether its verifier-admitted outputs can also become training supervision.

The interpretation depends on target provenance. A verifier can select a valid
candidate without creating it. A procedural generator may already contain the
answer. A reasoning regime may expose behavior that a cheaper serving regime
rarely produces. We therefore distinguish oracle-target SFT,
same-checkpoint verifier-selected self-distillation, and stronger-teacher
verifier-selected distillation.

Our primary result uses the external \planbench{} Blocksworld replanning
surface and pinned \valsys{} plan validation
\citep{valmeekam2023planbench,kclplanningval}. The domain, task semantics, and
verifier were not authored for this project. We generated fresh cases,
excluded structural overlap with published and prior experimental problems,
kept harvest and evaluation partitions disjoint, and never serialized an
oracle replacement plan into a target artifact.

On this surface, \qwenmodel{} thinking generated 24 verifier-admitted plans
from 32 harvest cases. A LoRA trained the same checkpoint for non-thinking
execution. On 80 unopened cases, base non-thinking, base thinking, and the
trained non-thinking adapter reached 1, 30, and 57 verifier-accepted plans.
Against the cheap base, the adapter gained 56 paired cases with zero
regressions and improved both the 5-block and 6-block strata. It was
schema-valid on all 80 cases and used approximately 1/56 of thinking's mean
latency and 1/86 of its completion tokens. The 57-versus-30 comparison is
operationally important but descriptive because adapter superiority over
thinking was not preregistered.

A prospective replication tests whether that primary result depended on a
lucky LoRA seed. It holds the 24 targets, model revision, training recipe, and
80-case corpus fixed, varies eight new training seeds, and evaluates each
adapter and its matched base under three inference realizations. Every seed
produces a clear positive lift, with adapter results from 45/80 to 70/80 against
1/80 for every base report. All 24 adapter reports pass the semantic and
interface gates, every report improves both difficulty strata, and no report
contains a base-only semantic regression. The exact training-seed sign-flip
test gives $p=0.0078125$. This is now the paper's strongest result.

A matched target-selection ablation isolates a previously unresolved part of
that pipeline. From the same frozen pool of 52 schema-valid Qwen candidates, we
train three 24-target adapters selected by \valsys{}, blinded Qwen
self-selection, or schema validity alone. At the initial training seed, the
three adapters reach 102, 69, and 55 accepted plans on 160 new cases,
respectively, while base reaches 1. Against model self-selection, verifier
selection gains 33 paired cases and improves both block-count strata.

Prospective seed studies materially narrow that finding. Across eight new
training seeds and three inference realizations per seed, the \valsys{} versus
model-self contrast has two clear positive seeds, one clear reverse seed, and
five null seeds; its exact seed-level test gives $p=0.3672$. A matched
\valsys{} versus schema-only study has four clear positive and four null seeds,
but fails its preregistered reliability gate with $p=0.0703$. The stable claim
is therefore not that semantic selection reliably improves downstream
capability. It is that the verifier deterministically establishes which targets
satisfy the declared contract and preserves that admission provenance, while
the capability consequence of selection is training-seed sensitive under this
low-data recipe.

A prospectively frozen companion arm changes the provenance. Phi-4-reasoning
generated 24 plans that \valsys{} selected before they trained base Phi-4. Base,
teacher, and distilled student reached 2, 30, and 51 verifier-accepted plans.
The student was schema-valid on all 80 cases, compared with 35 for base and 41
for the reasoning teacher. This is stronger-teacher distillation, not
teacher-free self-training. The Qwen arm remains the load-bearing teacher-free
evidence on the external surface. The synthetic program supplies a separate
same-family Phi self-training replication.

The earlier synthetic program supplies the development chain behind this
external result. Oracle-target experiments establish teachability. Model-
generated targets then produce cumulative gains in two workflow constructions,
a construction-separated replication in a third, a fixed-benchmark Phi
self-training replication, selective transfer, and frozen stops when another
round adds no measured information. These experiments remain important, but
they no longer carry the paper's external-validity conclusion.

Our contributions are:

\begin{itemize}
  \item a provenance-explicit record separating the contract, failed attempt,
  verifier decision, accepted target, target producer, and training recipient;
  \item external same-checkpoint self-distillation in which independently
  verified Qwen thinking outputs train cheap non-thinking execution without an
  oracle or stronger teacher, with the primary lift replicated across eight
  prospective LoRA training seeds;
  \item matched and prospectively replicated target-selection studies showing
  that the original large verifier-selection advantage is conditional on
  training seed rather than a reliable capability law;
  \item a provenance-separated Phi distillation arm showing that the same
  pipeline can convert a stronger reasoning model's accepted outputs into a
  reliable base-model specialist; and
  \item synthetic evidence for teachability, cumulative learning,
  construction robustness, frontier movement, selective transfer, and
  measurement-specific stopping.
\end{itemize}

Algorithmically, the method is filtered supervised training adjacent to
offline reward-filtered self-training, rejection-sampling fine-tuning, and
reinforcement learning for reasoning
\citep{gulcehre2023rest,yuan2023scaling,shao2024deepseekmath,deepseekai2025r1};
we do not claim a new optimizer. The central claim is narrower: within a
machine-verifiable task contract, verifier-admitted model outputs can become
supervision that consolidates occasionally demonstrated or expensive
capability into reliable one-shot execution. Qwen establishes this without
oracle labels or a stronger teacher and now across eight prospective training
seeds. Phi establishes the complementary stronger-teacher route. The external
evidence remains one synthetic planning domain, one tested difficulty band,
one fixed Qwen target set and recipe, and one Phi training seed, not
independently authored institutional work.

%% file: sections/02_problem.tex
\section{Problem Formulation}

\subsection{Verifiable structured repair}

Let a public contract $C$ define required tasks, allowed actions, dependencies,
state transitions, resource constraints, or completion conditions. A candidate
plan $y$ is evaluated by a deterministic verifier
$V(C,y)$, which returns a transport verdict, a semantic pass/fail verdict, and a
set of failed checks:

\begin{equation}
  V(C,y) = (t, p, F), \qquad F=\{f_1,\ldots,f_k\}.
\end{equation}

A repair is successful when the model receives the public contract and task
state, optionally including a malformed prior plan $y^{-}$, then emits
$\hat{y}$ such that $V(C,\hat{y})$ is transport-valid and $F=\emptyset$.
Evaluation permits one model response and no retry. This makes adaptation,
rather than inference-time search depth, the measured treatment.

The verifier defines correctness only within its declared contract. It does not
establish that the contract is complete for real institutional work. We use
\emph{verifier-confirmed} rather than objective correctness throughout.

\subsection{Three supervision surfaces}

The experiments distinguish three inputs that are often conflated:

\begin{enumerate}
  \item \textbf{Verdict:} the prior plan is invalid.
  \item \textbf{Localized governance trace:} exact failed check identifiers,
  rule references, and observed violations.
  \item \textbf{Verified target:} a complete plan $y^{+}$ for which
  $V(C,y^{+})$ passes.
\end{enumerate}

Localized feedback may reduce the search problem by naming the residual.
Correct targets may instead teach the full output distribution directly. The
primary LoRA treatment contains both, so a matched ablation is required to
attribute their roles.

\subsection{Target origin}

Target validity and target origin are separate variables. We use the following
taxonomy:

\begin{description}
  \item[Procedural oracle target.] Constructed by a generator or author with
  direct access to the valid schedule, then checked by the verifier.
  \item[Verifier-selected target.] Generated without target access and admitted
  only after passing the independent verifier.
  \item[Same-checkpoint self-distillation target.] Generated by one inference
  regime and used to train another serving regime of the same checkpoint.
  \item[Stronger-teacher distillation target.] Generated by a separately
  trained teacher and used to train a distinct student checkpoint.
  \item[Diagnosis-steered target.] Generated through a sequence in which each
  failed candidate changes the next prompt through a new verifier residual.
  \item[Human-adjudicated target.] Corrected or approved by an authorized human
  under a recorded institutional policy.
\end{description}

The synthetic precursor uses procedural oracle targets. Later target-hidden
workflow studies train Qwen or Phi on their own verifier-selected outputs. The
external Qwen study uses same-checkpoint self-distillation from thinking to
non-thinking execution. The external Phi study uses a separately trained
reasoning teacher. This distinction controls which results support a
teacher-free claim and which support only distillation.

\subsection{Research questions}

\begin{description}
  \item[RQ1: External teacher-free consolidation.] Can thinking outputs from a
  checkpoint, admitted only by an external verifier, train the same checkpoint
  for more reliable non-thinking execution on fresh external cases?
  \item[RQ2: Selection policy.] When the candidate pool and training exposure
  are fixed, does independent semantic verification reliably select targets
  that produce greater downstream capability than blinded model self-selection
  or schema-only selection across training seeds?
  \item[RQ3: Provenance pathway.] Does the same governance-selected pipeline
  also work when the target producer is a separate stronger reasoning model?
  \item[RQ4: Synthetic mechanism.] Do oracle-target teachability, model-
  generated target sourcing, and cumulative learning reproduce under controlled
  workflow contracts?
  \item[RQ5: Boundaries.] Where do iteration, search reachability,
  cross-construction transfer, interface reliability, and frozen continuation
  gates stop supporting broader claims?
\end{description}

RQ1 is positive for external semantic consolidation across eight prospective
Qwen training seeds and negative for the separate paired interface-cure
endpoint in the parent experiment. RQ2 is negative under the preregistered
seed-level reliability tests: both selector comparisons are heterogeneous, and
the verifier-versus-model contrast includes a clear reverse seed. RQ3 is
positive under the stronger-teacher classification. RQ4 is supported across
the oracle precursor and three self-authored constructions. RQ5 is bounded by
selective transfer, saturated measurements, non-material later rounds, one
external planning band, fixed Qwen target partitions and recipes, one Phi seed,
and the absence of institutional work.

%% file: sections/03_method.tex
\section{Governance Records as Supervision}

\subsection{The governance record}

Each retained record links five objects:

\begin{equation}
  R = (C,\ y,\ V(C,y),\ o,\ g),
\end{equation}

where $C$ is the declared contract, $y$ is the model output, $V(C,y)$ is the
verifier result, $o$ records target origin and model lineage, and $g$ records
the applicable experimental gate. A passing output may become a training
target only when its origin is admissible for the stated claim. Failed outputs
remain part of the provenance trail but cannot enter training as accepted
targets.

The verifier is load-bearing for admission, but it does not author a candidate
or establish that the contract captures institutional truth. Training performs
the consolidation. The target-selection studies measure whether semantic
admission also produces a downstream capability advantage relative to model
self-selection and schema-only selection. Their multiseed results do not
establish such an advantage reliably.

\subsection{Synthetic workflow surface}

Each synthetic case defines a workflow vocabulary rather than a natural-language
answer key. The public contract declares tasks and fields sufficient for the
verifier to check:

\begin{itemize}
  \item required-task coverage and uniqueness;
  \item dependency ordering;
  \item allowed tool assignment;
  \item resource-capacity limits;
  \item execution windows; and
  \item same-batch and not-same-batch relations.
\end{itemize}

The generator first constructs a valid canonical schedule and then derives a
contract that admits it. Training and evaluation rows are created by applying
one of eight single or four compound mutation families to a copy of that
schedule. The unchanged canonical schedule becomes the supervised assistant
target. The verifier evaluates both the malformed prior and the target, and the
row records the target's provenance.

This ordering matters. The verifier is an independent acceptance function at
evaluation, but it is not the source of the precursor training labels. The
label exists before verification. These studies establish teachability and
signal attribution, not oracle-free learning.

\subsection{Synthetic example construction}

A localized-trace example contains:

\begin{equation}
  x_{\mathrm{local}} =
  [C,\ y^{-},\ \mathrm{id}(F),\ \mathrm{rules}(F),\ \mathrm{observations}(F)],
\end{equation}

with target $y^{+}$. A verdict-only example holds $C$, $y^{-}$, and $y^{+}$
fixed but replaces the detailed residual with a generic failure statement:

\begin{equation}
  x_{\mathrm{verdict}} = [C,\ y^{-},\ \text{``the plan is invalid; repair it''}].
\end{equation}

The prompt never contains the target. Dataset audits check this condition
before training. Every evaluation cell receives the same public contract,
malformed plan, output schema, and verifier. The primary Qwen ablation changes
only the training feedback surface and then holds localized inference feedback
fixed.

\subsection{External planning surface}

The external study uses \planbench{} Blocksworld repository task T6,
replanning after an execution-time state change. The repository labels this
surface T6 even though replanning is the seventh item in the paper's one-based
curriculum list \citep{valmeekam2023planbench,planbenchrepo}. \planbench{}
supplies the externally authored domain and instance generator. The pinned
\valsys{} implementation supplies executable plan validation
\citep{kclplanningval}. The study pipeline supplies fresh-case construction, structural
contamination checks, target selection, provenance, training orchestration,
and evaluation.

The formal corpus contains 32 harvest cases and 80 evaluation cases, evenly
split between 5 and 6 blocks. Raw, normalized, and object-renaming-invariant
fingerprints exclude overlap with 2,207 published PlanBench problems, prior
experimental PlanBench corpora, and another formal partition. Oracle replacement
plans are not serialized into case or target artifacts.

For Qwen, thinking mode receives up to eight independent attempts per harvest
case. A syntax-only canonicalizer may add PDDL parentheses around exact
declared action and object sequences, but cannot add, remove, reorder, or repair
actions. \valsys{} evaluates every schema-valid plan. The first accepted plan
from each successful case becomes a target. The Phi arm uses the same corpus,
prompts, verifier, partitions, and gates, but a delimiter-only extractor admits
only the final section emitted after the reasoning trace.

\subsection{Parameter-efficient adaptation}

We use low-rank adaptation, which freezes base-model weights and trains small
rank-decomposition modules \citep{hu2022lora}. The external Qwen and Phi
adapters each train on exactly 24 verifier-selected plans for three fixed
epochs in BF16 on one NVIDIA A100-SXM4-80GB. Both use rank 16, alpha 32,
dropout 0.05, learning rate $10^{-4}$, 5\% warmup, cosine decay, and fixed
exposure. Final checkpoints are selected without evaluation-driven tuning.

The synthetic precursor and workflow self-training studies use the same broad
LoRA approach under their separately frozen protocols. Full hyperparameters,
partition rules, and amendments are reported in the appendices rather than the
main argument.

\subsection{Matched target-selection ablation}

The selection ablation reuses the frozen Qwen harvest pool: 24 successful
cases containing 52 schema-valid candidates. Thirteen cases have one candidate
and 11 are contested. Three selectors each produce exactly one target per case:
\valsys{} admits the unique semantically valid candidate, Qwen thinking chooses
an opaque candidate ID without verifier information, and a schema-only rule
takes the first candidate in a frozen opaque shuffle. The model selector cannot
rewrite, repair, or combine candidates.

The original three 24-target arms fix the case IDs, candidate pool, prompt
surface, checkpoint, LoRA recipe, training seed, three epochs, and 72 optimizer
steps. Checkpoint selection uses fixed exposure rather than evaluation. The
resulting non-thinking adapters and the unadapted base are evaluated once on
160 newly generated, structurally excluded cases, evenly divided between 5 and
6 blocks. The frozen primary comparison is paired \valsys{} acceptance for
verifier selection against model self-selection. It requires at least eight net
gains, exact two-sided $p<0.05$, and a positive effect in both strata.

\subsection{Prospective training-seed replications}

Two prospective studies separate robustness of the intervention from
robustness of its target-selection explanation. The primary Self-24 replication
holds the original 24 verifier-admitted Qwen targets, base revision, LoRA
recipe, and 80-case evaluation corpus fixed. It trains eight new adapters under
seeds 2026113002 through 2026113009 and evaluates each adapter and a matched
base under three frozen inference-seed bases. The training seed is the
inferential unit. A clear seed requires at least two of three realizations to
pass the semantic gate, no realization with a negative net, and at least two
realizations to pass the 95\% schema-validity and zero-endpoint-error interface
gate.

The selector replication reuses the byte-identical \valsys{} and model-self
24-target partitions, checkpoint, 72-step recipe, and 160-case corpus. It
trains both arms under seeds 2026120305 through 2026120312 and evaluates each
under three inference realizations. A separate matched study prospectively
trains schema-only adapters under the same eight seeds and compares them with
the archived \valsys{} arms. Each study uses an exact two-sided sign-flip test
over eight seed-level paired effects. The \valsys{} versus model-self
study requires a consistent positive direction under its frozen rule. The
\valsys{} versus schema study requires at least six of eight clear positive
seeds and seed-level $p<0.05$. Repeated case outcomes describe magnitude but
are not treated as independent observations.

\subsection{Verifier-selected target records}

For target-hidden studies, the model receives the public contract and output
schema, plus a malformed plan or changed planning state when the task provides
one, but no target plan. Each stochastic candidate is retained with its prompt
hash, model and adapter identity, serving regime, sampling configuration,
verifier verdict, token counts, and attempt index. The first verifier-passing
candidate within a frozen per-case budget becomes an admissible target.
Rejected predecessors are preserved rather than discarded.

This record distinguishes three effects. Search yield asks whether valid
behavior appears at all within the budget. One-shot evaluation asks whether
training makes that behavior immediate on fresh cases. Cumulative training asks
whether an adapted model can harvest targets that support another adapter. A
new round is permitted only by a predeclared gate on a case-disjoint partition.
No unresolved row is backfilled with a procedural target under a self-sourcing
label.

\subsection{Admission and scoring}
\label{sec:admission-scoring}

A test response passes only when it is parseable under the frozen output
contract and the deterministic workflow verifier returns no failed checks.
Transport validity is reported separately so semantic gains cannot be reduced
to formatting repair. Primary measures are:

\begin{itemize}
  \item one-shot verifier-confirmed pass rate;
  \item paired both-pass, adapter-only, base-only, and both-fail transitions;
  \item result by single and compound mutation family;
  \item transport validity;
  \item runtime and output tokens; and
  \item predeclared gate outcomes.
\end{itemize}

Target-hidden search additionally reports first-attempt yield, fixed-budget
coverage, paired-only harvests, unique parsed plans, calls, and tokens. The
appendices retain the conventional pass@N notation where needed. Any
fixed-budget result is a bounded empirical search measure, not proof of
absolute model reachability.

The synthetic rows are clustered within generated cases. The external
evaluations contain distinct cases but still represent fixed generated
benchmark samples. We report exact paired counts because every plan is
separately verified. For the prospective studies, the training seed is the unit
of inference and inference realizations are repeated measurements. These tests
estimate training-seed robustness for fixed targets, recipes, and corpora, not
robustness to new target harvests, benchmark samples, models, or domains.

\subsection{Why retain the verifier after training?}

The adapter is not promoted to an authority. It proposes a repair; the verifier
still decides whether the declared contract is satisfied. Training aims to
increase first-pass yield and reduce repeated inference, not to replace
independent completion checks. In a deployment, an invalid response remains a
retry, rerouting, or escalation event and can become a future governed training
example only after its correction and authority are established.

%% file: sections/04_core_evidence.tex
\section{Core Evidence}
\label{sec:core-evidence}

\subsection{External design and provenance}

The external experiment changes the task and verifier authorship relative to
the synthetic development program. It uses fresh \planbench{} Blocksworld
replanning cases and pinned \valsys{} validation. The formal corpus was
generated after the protocol freeze and contains 32 harvest cases and 80
evaluation cases, evenly split between 5 and 6 blocks. The partitions have zero
case and structural overlap. Neither partition overlaps the indexed published
PlanBench problems or prior experimental PlanBench corpora, and no oracle replacement
plan appears in a target artifact.

Both model paths use the same cases, prompts, syntax contract, verifier, fixed
training exposure, and semantic gates. Their target provenance differs, as
Table~\ref{tab:external-provenance} makes explicit. The Qwen path is the primary
teacher-free result. The Phi path is complementary stronger-teacher
distillation and cannot inherit the Qwen classification.

\begin{table}[tbp]
\centering
\footnotesize
\begin{tabular}{@{}p{0.09\linewidth}p{0.20\linewidth}p{0.20\linewidth}p{0.18\linewidth}p{0.20\linewidth}@{}}
\toprule
Path & Target producer & Training recipient & Target provenance & Frozen reading \\
\midrule
Qwen & Qwen3-14B thinking & Same checkpoint, non-thinking & No oracle; no stronger teacher & Primary lift replicated \\
Phi & Phi-4-reasoning & Separate base Phi-4 & No oracle; stronger teacher & Full-band distillation \\
\bottomrule
\end{tabular}
\caption{The two external paths share the governance-selected pipeline but not
the same causal classification.}
\label{tab:external-provenance}
\end{table}

\subsection{Same-checkpoint Qwen self-distillation}

Qwen thinking produced a verifier-admitted plan for 24 of 32 harvest cases
within eight attempts. Thirteen cases passed on the first attempt, leaving 11
cases that exposed the occasional-versus-immediate capability gap required for
consolidation. Every selected plan was individually accepted by \valsys{}, both
block-count strata contributed targets, and the training partition contained
24 distinct plan hashes. A rank-16 LoRA then trained the same checkpoint for
non-thinking execution in 72 fixed optimizer steps.

On the 80 unopened cases, base non-thinking, base thinking, and Self-24
non-thinking reached 1, 30, and 57 \valsys{}-accepted plans. Relative to base
non-thinking, Self-24 produced 56 adapter-only successes, zero base-only
regressions, one both-pass case, and 23 both-fail cases. The exact two-sided
paired probability is below $10^{-10}$. Gains occur in both prespecified
strata: 5-block performance moves from 1/40 to 30/40, and 6-block performance
moves from 0/40 to 27/40.

\begin{figure}[tbp]
\centering
\includegraphics[width=\linewidth]{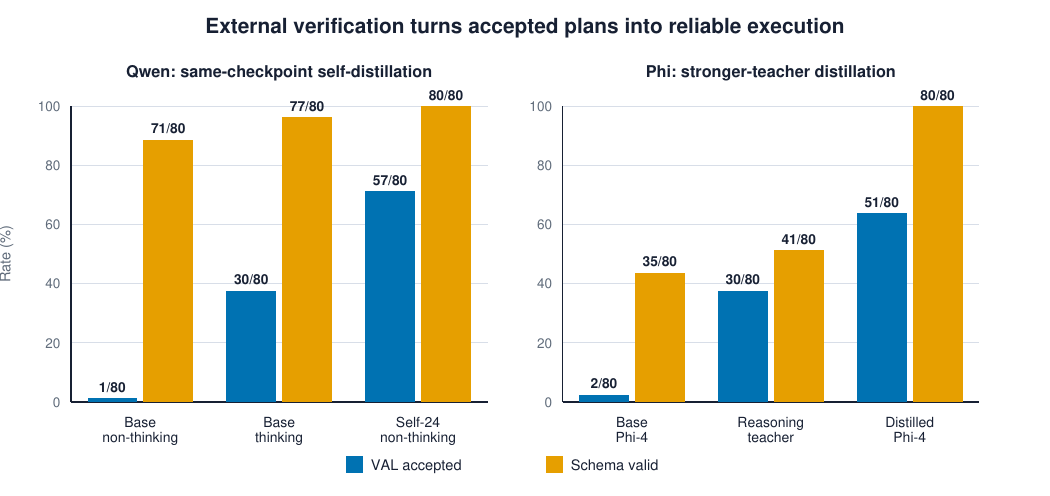}
\caption{Fresh external outcomes on the shared 80-case PlanBench evaluation.
The Qwen path is teacher-free same-checkpoint self-distillation. The Phi path
uses a separate stronger reasoning teacher. Exact counts are printed above the
bars.}
\label{fig:external-outcomes}
\end{figure}

Figure~\ref{fig:external-outcomes} shows the semantic and interface outcomes
for both external paths. The Qwen adapter is schema-valid on 80/80 cases, with
zero endpoint errors. Its frozen result is nevertheless
\texttt{SEMANTIC\_ONLY}, because the co-primary interface endpoint tested
paired improvement over thinking rather than absolute deployment validity.
Thinking was already schema-valid on 77/80 cases. Self-24 added only three
paired schema gains, with no losses and $p=0.25$, below the frozen five-gain and
$p<0.05$ requirements. This negative endpoint means that an interface cure was
not demonstrated. It does not mean the deployed adapter has unresolved schema
failures.

The adapter also exceeds thinking descriptively, 57/80 versus 30/80. This was
not a preregistered superiority test. Its defensible use is operational: the
filtered specialist produces verifier-accepted plans more consistently on this
screen than a fresh draw from the expensive reasoning regime that supplied its
targets.

\FloatBarrier
\subsection{The primary lift replicates across training seeds}

The prospective Self-24 replication holds the original 24 thinking-generated,
\valsys{}-admitted targets, Qwen revision, 72-step LoRA recipe, and 80-case
evaluation corpus fixed. Eight new training seeds each produce one adapter,
evaluated with a matched regenerated base under three frozen inference
realizations. The parent adapter is rerun only as context and is excluded from
the prospective decision.

All eight seeds are clear positive lifts under the frozen rule. No seed is null
or reverse. The exact sign-flip test over the eight training-seed effects gives
$p=0.0078125$, the smallest attainable two-sided value with eight nonzero
effects sharing one sign. Across the 24 adapter reports, accepted plans range
from 45/80 to 70/80, with mean 58/80; every matched base report reaches 1/80.
Every replicate improves both block-count strata and has zero base-only
semantic regressions. Adapter schema validity is 1910/1920 (99.48\%), compared
with 1662/1920 (86.56\%) for base, and every replicate is endpoint-clean.

\begin{figure}[tbp]
\centering
\includegraphics[width=\linewidth]{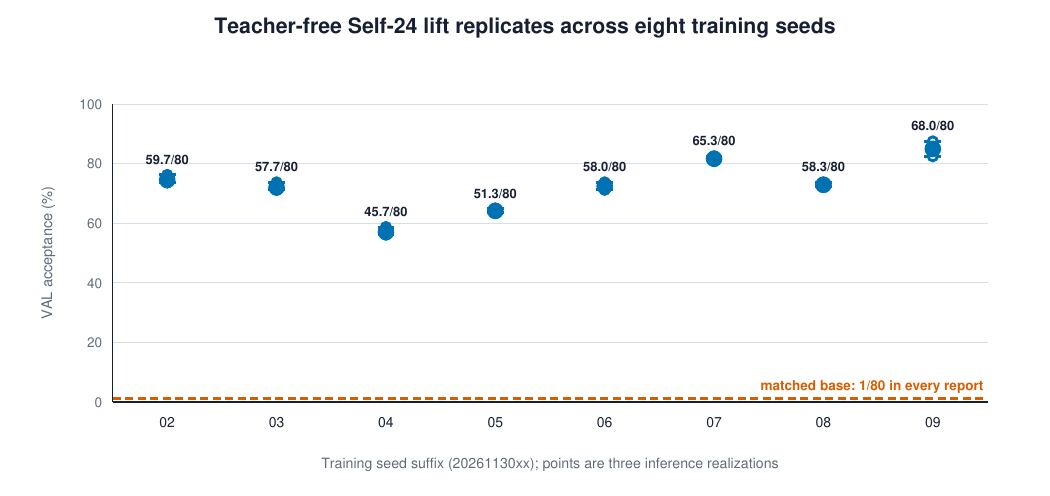}
\caption{Prospective replication of the primary teacher-free Self-24
intervention. Filled points are per-seed means and open points are three frozen
inference realizations. The base result is 1/80 in every matched report. The
training seed, not the repeated case, is the inferential unit.}
\label{fig:self24-replication}
\end{figure}

Figure~\ref{fig:self24-replication} shows the three inference realizations for
each training seed. The small within-seed ranges and uniformly positive
between-arm separation establish that the original intervention was not a
lucky LoRA initialization under this fixed target set and recipe. They do not
test a new target harvest, evaluation corpus, model revision, or training
recipe.

\FloatBarrier
\subsection{Selection performance is training-seed sensitive}

The matched ablation asks whether verifier selection matters once candidate
generation and training exposure are fixed. The source pool contains the same
24 harvest cases and 52 schema-valid Qwen candidates used by the teacher-free
path. Thirteen cases are singletons and 11 require a choice among candidates.
The \valsys{} selector supplies 24/24 semantically valid targets. Blinded Qwen
self-selection supplies 20/24, and schema-only selection supplies 15/24.

At the initial training seed, base non-thinking reaches 1/160 \valsys{}
acceptance, while schema-selected, model-self-selected, and
\valsys{}-selected adapters reach 55/160, 69/160, and 102/160. Against model
self-selection, verifier selection records 41 verifier-only successes, eight
self-only successes, paired net $+33$, and exact
$p=1.9647\times10^{-6}$. Both strata are positive. The frozen result is valid
for that matched seed, but the prospective studies show that it is not a
characteristic selector effect.

Across eight new training seeds, the \valsys{} versus model-self study has two
clear \valsys{} seeds, one clear model-self seed, and five seeds with no
detectable effect. The seed effects, summed over three inference realizations,
are $[+67,+21,+40,-33,+29,-66,+7,+59]$. The exact seed-level sign-flip test
gives $p=0.3672$, so the frozen reading is
\texttt{NO\_REPLICATED\_SELECTOR\_ADVANTAGE}. Strong positive and negative
seeds are each internally stable across their three inference realizations.

The matched \valsys{} versus schema-only study produces four clear positive
seeds and four null seeds, with no clear schema reverse. Its seed effects are
$[+88,-23,+159,-29,+116,+3,+39,+96]$. Six of eight are numerically positive,
but only four clear the per-seed gate and the exact sign-flip value is
$p=0.0703$. It therefore returns
\texttt{NO\_REPLICATED\_SEMANTIC\_SELECTION\_ADVANTAGE}. This is not an
equivalence result and does not show that verifier-rejected targets are
harmless.

\begin{figure}[tbp]
\centering
\includegraphics[width=\linewidth]{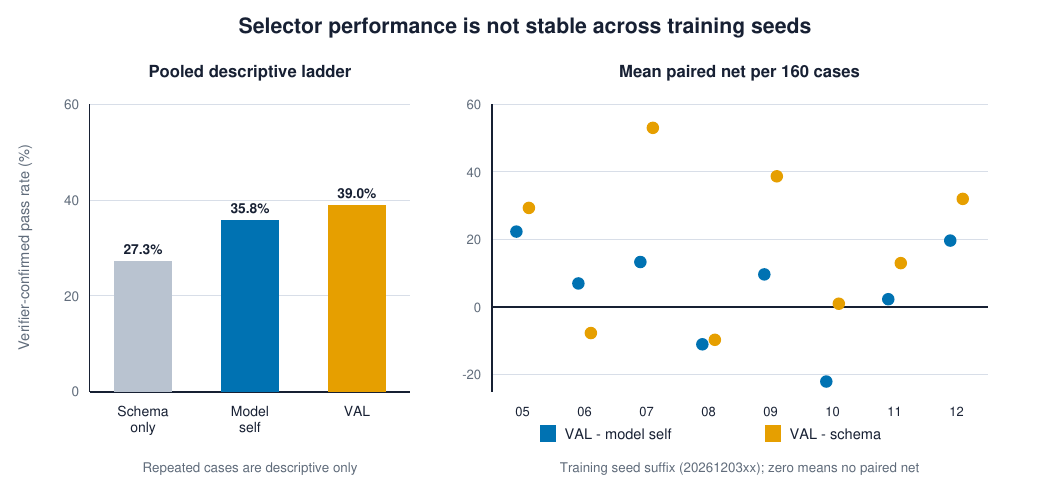}
\caption{Selector outcomes across eight prospective training seeds. The left
panel is a descriptive pooled ladder. The right panel shows each seed's mean
paired net across three inference realizations; both sets of seed-level
contrasts cross zero and neither passes its preregistered seed-level reliability
test. The original base result on this 160-case corpus was 1/160, measured at
one training seed and one inference realization; it is excluded from the common
eight-seed ladder.}
\label{fig:selector-robustness}
\end{figure}

Figure~\ref{fig:selector-robustness} shows both readings. Pooled descriptively,
schema-only, model-self, and \valsys{} adapters reach 1048/3840 (27.3\%),
1373/3840 (35.8\%), and 1497/3840 (39.0\%). The monotone aggregate cannot
override the seed-level tests because the 160 cases are repeated across arms,
realizations, and training seeds. The stable contribution is deterministic
admission: only the \valsys{} partition carries 24/24 semantic receipts. The
incremental downstream capability benefit of that admission policy is not
reliably established under this low-data LoRA recipe.

\FloatBarrier
\subsection{Stronger-teacher Phi distillation}

The Phi protocol was frozen before the formal Qwen response partitions were
opened. Phi-4-reasoning then generated a separate set of candidates on the same
32 harvest cases. A delimiter-only extractor admitted only the final answer
after the reasoning section; incomplete reasoning traces were rejected rather
than mined. \valsys{} admitted 24 distinct targets, with first-attempt success
on 10 cases and fixed-budget success on 24. Those plans trained base Phi-4
under the same fixed LoRA recipe.

On the shared 80-case evaluation, base Phi-4, Phi-4-reasoning, and the distilled
adapter reached 2, 30, and 51 verifier-accepted plans. Relative to base, the
adapter gained 49 paired cases with zero regressions and exact $p<10^{-10}$.
The 5-block stratum moves from 2/40 to 23/40, and the 6-block stratum moves from
0/40 to 28/40.

The Phi interface result is also positive under its frozen gate. Schema
validity moves from 35/80 for base and 41/80 for the reasoning teacher to 80/80
for the distilled adapter. Against the teacher, this is 39 paired gains, zero
losses, and exact $p<10^{-10}$. The reliability was therefore not inherited
from a consistently formatted teacher. The supported causal statement is that
the verifier-admitted training pipeline produced interface reliability absent
from ordinary teacher outputs. The Qwen selector studies do not decompose the
Phi interface gain or give the verifier sole credit for the full effect, and
the Phi arm has no matched multiseed selection ablation.

\begin{figure}[tbp]
\centering
\includegraphics[width=\linewidth]{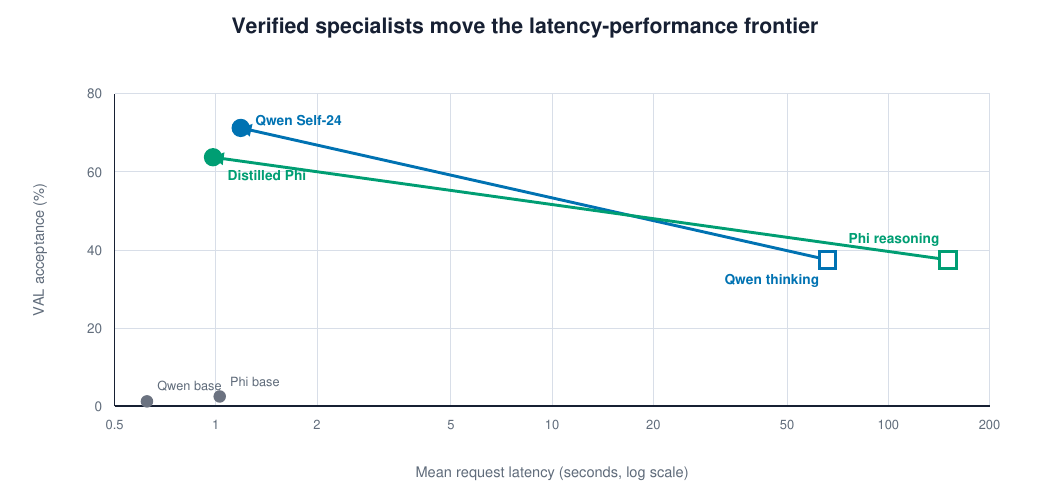}
\caption{Verified performance versus mean request latency on the external
evaluation. Arrows connect the expensive target-producing regime to the cheap
trained specialist. The source-versus-specialist semantic comparisons are
descriptive compression diagnostics, not preregistered superiority tests.}
\label{fig:external-efficiency}
\end{figure}

As with Qwen, the student's 51/80 versus the teacher's 30/80 is descriptive,
not a preregistered general teacher-superiority claim. Figure
\ref{fig:external-efficiency} shows why the comparison remains operationally
important. The trained specialists occupy a different cost-performance region
from fresh reasoning: Qwen runs at approximately 1/56 of thinking's mean
latency and 1/86 of its completion tokens; Phi runs at approximately 1/153 of
the reasoning teacher's latency and 1/215 of its completion tokens.

\FloatBarrier
\subsection{The synthetic program supplies mechanism and boundary evidence}

The external result did not begin from an untested recipe. Oracle-target
precursors first established that low-rank adaptation can teach structured
workflow repair in Phi-4 and Qwen3-14B, and a matched ablation showed that
complete accepted targets carry most of the measured teaching signal. Because
the procedural generators created those schedules before verification, the
precursors motivate the target-hidden program but do not support oracle-free
learning.

The later synthetic experiments remove target access and train only on model-
generated plans admitted by the frozen workflow verifier. Table
\ref{tab:core-self-sourced} summarizes the one-response evidence. Qwen produces
material gains in Generators A and B across cumulative rounds. Generator C
provides a construction-separated first-round replication with a higher base
rate and then saturates its frozen measurement. A separate Phi-4 experiment
reproduces one complete self-training round on the frozen synthetic Generator B
benchmark.

\FloatBarrier
\begin{table}[tbp]
\centering
\small
\begin{tabular}{lrrrr}
\toprule
Construction and partition & Base & First adapter & Second adapter & Later adapter \\
\midrule
Generator A, initial & 10/48 & 21/48 & n/a & n/a \\
Generator A, later & 9/48 & 21/48 & 33/48 & 35/48 \\
Generator B, Qwen & 12/80 & 44/80 & 62/80 & n/a \\
Generator C, Qwen & 57/128 & 128/128 & 126/128 & n/a \\
Generator B, Phi self-training & 10/80 & 77/80 & n/a & n/a \\
\bottomrule
\end{tabular}
\caption{Fresh synthetic one-response repair after training only on model-
generated, verifier-selected targets. The two Generator A rows use different
case-disjoint evaluation partitions. The later Generator A step does not clear
its frozen materiality gate.}
\label{tab:core-self-sourced}
\end{table}

\begin{figure}[tbp]
\centering
\includegraphics[width=\linewidth]{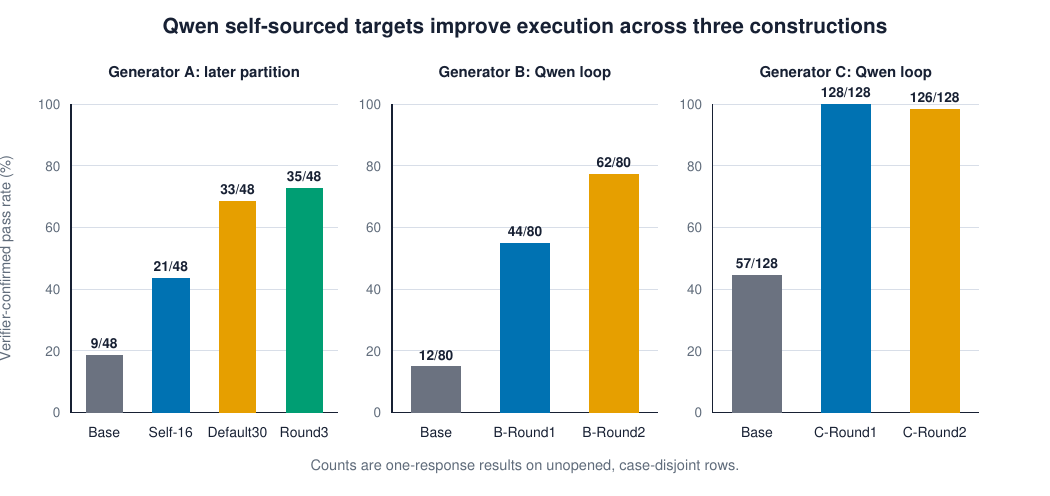}
\caption{Qwen self-sourced one-response learning across three synthetic
constructions. Generator C's high base rate and first-round saturation limit
its evidence about iteration.}
\label{fig:iterative-learning}
\end{figure}

The detailed synthetic record matters because it prevents the external result
from being narrated as open-ended improvement. Generator A supports two
material cumulative gains and then a non-material third step. Generator B
supports two one-response gains, but its second fixed-budget frontier increment
is below the continuation threshold. Generator C reaches the ceiling of its
one-response and fixed-budget measurements after one round. Cross-generator
transfer is aggregate-positive but family-selective. These are measured stops
and transfer boundaries, not model ceilings.

The combined evidence supports a precise hierarchy. The external Qwen arm is
the primary no-oracle, no-stronger-teacher result. External Phi shows a second
operational path with stronger-teacher provenance. The synthetic program shows
that the broader mechanism reproduces, accumulates, transfers selectively, and
stops under frozen gates. Appendix~\ref{app:planbench} gives the complete
external protocol and its negative qualifications. Appendix
\ref{app:synthetic-program} consolidates the synthetic controls, supporting
results, nulls, amendments, and interpretation boundaries.

%% file: sections/08_discussion.tex
\section{Discussion}

\subsection{The primary result is replicated capability consolidation}

The strongest result is not that one LoRA can memorize accepted plans. It is
that the same Qwen checkpoint can expose occasional valid planning behavior
under expensive thinking, an independently authored verifier can admit those
outputs, and the same 24 outputs can train reliable non-thinking execution
across eight prospective LoRA seeds. Every seed is a clear lift, every
realization improves both strata, and none records a base-only semantic
regression. The intervention's magnitude varies, but its sign and frozen
classification do not.

This changes the paper's evidentiary position. The original external result was
a strong description of one trained adapter. The replication establishes
training-seed robustness for a fixed intervention. The earlier self-authored
generators establish a controlled mechanism and its iteration boundaries;
PlanBench and VAL show that it survives a task specification and verifier not
authored for this project. The remaining externality gap is institutional and
distributional, not merely another LoRA seed.

\subsection{Selection policy is not a stable capability law}

The original matched ablation remains internally valid. At seed 2026120301,
all three adapters see 24 targets from the same cases and pool, use the same
checkpoint and recipe, and differ only in candidate choice. Its 102/160 versus
69/160 result therefore identifies a target-selection effect for that trained
pair. It does not identify the typical direction or size of the effect across
training seeds.

The prospective evidence closes that distinction. The \valsys{} versus
model-self contrast ranges from large positive to large negative values, with
one clear reverse seed and seed-level $p=0.3672$. The \valsys{} versus
schema-only contrast has a suggestive positive aggregate and no clear reverse,
but only four of eight clear seeds and $p=0.0703$. Neither satisfies its frozen
reliability claim. In this 24-example LoRA regime, training seed can reverse the
performance conclusion supported by one otherwise matched experiment.

This negative result does not erase the verifier's governance role. The
\valsys{} partition contains 24/24 independently admitted targets, compared
with 20/24 for model self-selection and 15/24 for schema-only selection. Only
the first partition supports a deterministic statement that every training
target satisfied the declared contract. That is an admissibility and
provenance guarantee, not a replicated claim of downstream performance
superiority.

\subsection{Target provenance changes the claim}

Qwen and Phi implement the same high-level pipeline but answer different
scientific questions. Qwen uses one checkpoint in two serving regimes, so its
target record supports teacher-free same-checkpoint self-distillation. Phi uses
a separately trained reasoning teacher and base student, so its result is
governance-selected distillation. Combining them without this distinction
would overstate the Phi evidence and understate the Qwen evidence.

The Phi interface outcome is especially informative. The teacher is
schema-valid on only 41 of 80 cases, while the student reaches 80 of 80. The
student therefore did not inherit reliability from a consistently formatted
teacher. The verifier-admitted training pipeline converted a noisy source of
occasional valid plans into a reliable specialist. The Qwen studies show that
selection policy can matter substantially but do not establish a stable
direction across training seeds. They also do not identify every interaction
among target content, repeated exposure, ordinary SFT, and verifier admission
in the Phi arm.

\subsection{The specialist can descriptively exceed its source}

On this frozen screen, Qwen Self-24 reaches 57 verifier-accepted plans versus
30 for thinking, and distilled Phi reaches 51 versus 30 for its reasoning
teacher. Neither source-versus-student comparison was preregistered as a
superiority test. They do not show that a specialist became generally more
capable than its source.

The narrower explanation is consistent with filtered consolidation. Expensive
reasoning generates a noisy distribution containing occasional successes. The
verifier selects successes from multiple cases. Supervised training then shifts
one-response probability toward those accepted behaviors. A narrow specialist
can consequently execute the tested contract more consistently than a fresh
sample from the broader reasoning regime that generated its targets.

\subsection{Synthetic iteration remains scientifically useful}

The synthetic program contributes evidence that one external experiment alone
cannot provide. It separates procedural labels from self-sourced labels,
demonstrates two cumulative Qwen gains in two constructions, shows a
construction-separated replication, records frontier movement, reveals
family-selective transfer, and stops when frozen continuation gates fail. It
also includes a same-family Phi self-training replication without the stronger
teacher used in the external Phi arm.

These results constrain the mechanism. Verifier-selected training depends on
the model producing admissible fuel somewhere in its sampling distribution.
Training does not create targets where search finds none, and success under one
construction does not imply uniform transfer to another. Saturated evaluations
and non-material later rounds are reasons to change the measurement, not
evidence that the research program or model is exhausted.

\subsection{Operational interpretation and next experiments}

The measured architecture separates expensive acquisition from cheap repeated
execution. A reasoning regime can generate candidate work offline, governance
records can preserve the contract and verifier decision, and a narrow adapter
can serve routine cases while the verifier remains responsible for acceptance.
Failures remain retry, rerouting, or escalation events. Human resolutions
should become training data only after authority and independent completion are
recorded.

The largest external-validity gap is independently authored institutional work.
That partner-paced study is a separate validation track. Under direct
experimental control, the tractable next planning experiment should test the
specialist size floor rather than repeat a single large-to-small comparison.
The useful question is how small the deployed model can become while retaining
verifier-confirmed capability, interface reliability, and acceptable regression
and escalation rates. A later study can test whether a distilled model below
the original fuel band can then continue through self-training. These are
distinct validation, scale, and composition questions, not missing controls for
the present result.

%% file: sections/09_limitations.tex
\section{Limitations and Evidence Boundaries}

The external result remains one public synthetic planning domain. It covers
fresh PlanBench Blocksworld replanning instances with 5 or 6 blocks, not
harder bands, arbitrary planning, or independently authored enterprise work.
PlanBench and VAL change task and verifier authorship, while the study still owns
fresh-case construction, contamination checks, prompting, target selection,
training, and analysis.

The primary Qwen intervention is replicated across eight prospective training
seeds and three inference realizations per seed. This estimates seed robustness
for one fixed 24-target set, model revision, recipe, and 80-case corpus. It does
not test new target harvests, hyperparameters, evaluation samples, or model
families. The Phi arm remains one training seed, and Qwen and Phi share the same
80 cases. Additional external domains, target sets, larger case samples, and
general-capability regression batteries are required for broader claims.

The Qwen frozen reading is \texttt{SEMANTIC\_ONLY}. Its adapter is absolutely
reliable under the measured interface, with 80/80 schema-valid responses and
zero endpoint errors, but the preregistered paired interface-cure endpoint over
thinking did not pass. The small three-case schema gain and $p=0.25$ should not
be rewritten as a positive interface-cure result or as an adapter defect.

The descriptive student-versus-source comparisons were not preregistered
superiority tests. They support an operational compression reading on this
screen, not the general claim that trained specialists outperform reasoning
models. The Qwen-versus-Phi comparison is also confounded by model family,
checkpoint lineage, starting interface validity, generated targets, and
teacher behavior. It cannot isolate what a stronger teacher contributes.

The original matched Qwen ablation isolates target-selection policy at fixed
case IDs, candidate pool, target count, checkpoint, training recipe, and seed.
Its large \valsys{} advantage is conditional on that seed. Across eight new
seeds, neither \valsys{} versus model-self nor \valsys{} versus schema-only
passes its preregistered reliability test. These studies do not establish
equivalence, a universal target-purity law, or harmlessness of invalid targets.
They also reuse one candidate pool and evaluation corpus and do not decompose
every interaction among selected content, repeated exposure, ordinary SFT, and
semantic admission. The Phi interface effect has no matched multiseed selection
ablation.

Inference realization is not negligible in this serving stack. The prospective
studies use three realizations per adapter and treat them as repeated
measurements rather than independent training observations. One preliminary
selector evaluation was also withdrawn after the pinned verifier executable
failed to load its shared library and the evaluator mistakenly converted the
process failure into semantic rejection. The final reports rescore unchanged,
hash-sealed response bytes with fatal handling of unexpected verifier status.
This incident supports stronger measurement controls, not a model-performance
claim.

The synthetic program has separate boundaries. All three workflow generators
are self-authored. Their evaluation rows are clustered within generated cases,
Qwen is tested more extensively than Phi, and several reported frontiers use
fixed repeated-sampling budgets. Generator A transfer is family-selective.
Generator C begins with a high base rate and saturates the frozen measurements
after one round. These are valid controlled results, not institutional
prevalence estimates or model ceilings.

Verifier passage establishes compliance with the declared contract, not policy
legitimacy or real-world completion. An incomplete contract can admit the wrong
behavior perfectly. Deployment requires authority over contract changes,
versioned applicability, grounded facts, external completion evidence, appeal,
and human escalation. The experiments do not test those institutional
properties, routed multi-adapter composition, verifier replacement, or a
specialist size floor.

Finally, all runtime measurements come from research paths on one A100. The
reported latency and token ratios characterize these frozen arms, not
production throughput, concurrent adapter serving, queueing, or total cost of
target acquisition and human oversight.

%% file: sections/10_related_work.tex
\section{Related Work}

\subsection{Governed execution and provenance records}

Prior work defines governed execution as work whose decisions, completion, and
response to change remain supported by inspectable provenance
\citep{salas2026correct}. Its primary object is an auditable institutional
lifecycle over authority, facts, tasks, completion receipts, verdicts, and
invalidation. The record studied here is narrower: a declared task contract,
model attempt, verifier decision, accepted output, and target lineage. We test
a downstream use not evaluated in that work, namely whether the
verifier-admitted subset can train a bounded specialist. We do not infer that
provenance alone improves a model or extend the present evidence to decision or
change governance.

\subsection{Parameter-efficient specialization}

LoRA freezes pretrained weights and learns low-rank update matrices, enabling
task adaptation with substantially fewer trainable parameters
\citep{hu2022lora}. QLoRA extends this approach through quantized base weights
\citep{dettmers2023qlora}. Our focus is not a new adapter algorithm. We study
the provenance and causal role of machine-verifiable supervision supplied to
ordinary LoRA training.

\subsection{Self-training and bootstrapping}

STaR bootstraps rationales that lead to correct answers
\citep{zelikman2022star}; Self-Instruct generates and filters instruction data
\citep{wang2023selfinstruct}; ReST produces policy samples offline and uses a
reward signal to select data for improvement \citep{gulcehre2023rest}; and SPIN
iteratively trains a model against outputs from its prior policy
\citep{chen2024spin}. These approaches motivate learning from generated
behavior, while also making target provenance central.

Rejection-sampling fine-tuning collects model-generated correct reasoning paths
as augmented supervision \citep{yuan2023scaling}. This is the closest procedural
precedent for training on accepted model outputs. Our method changes the object
and record: a symbolic plan verifier admits targets under a pinned contract,
and the retained governance record binds each target to its producer, student,
and verifier receipt. The seed studies further show that better semantic target
admissibility need not produce a stable downstream capability advantage under
a low-data LoRA recipe.

Our oracle-target precursors are not self-training because their target plans
are procedurally constructed. The later synthetic Qwen and Phi programs are
filtered self-training: each model samples candidates, a deterministic verifier
admits passing outputs, and LoRA consolidates them on case-disjoint
evaluations. The external Qwen experiment is more specifically same-checkpoint
self-distillation, in which thinking supplies targets for non-thinking
execution. Frozen stopping rules keep these claims distinct from unrestricted
recursive improvement.

\subsection{Knowledge distillation and serving regimes}

Knowledge distillation transfers behavior from a teacher into a student
\citep{hinton2015distilling}. The external Phi arm belongs to this family, but
adds independently executed admission: only teacher plans accepted by VAL
become targets. Qwen differs because target producer and training recipient
share one checkpoint. Its treatment changes serving regime and adds a LoRA
rather than transferring from a separately trained teacher. This lineage
distinction is central to the no-stronger-teacher claim.

\subsection{Reinforcement learning with verifiable rewards}

Reinforcement learning with verifiable rewards uses mechanically checkable
outcomes to optimize a policy, prominently for mathematical and coding tasks.
DeepSeekMath introduces Group Relative Policy Optimization for mathematical
reasoning \citep{shao2024deepseekmath}, and DeepSeek-R1 demonstrates large-scale
reinforcement learning on verifiable tasks \citep{deepseekai2025r1}.

Our target-hidden loop uses the same scarce resource, a reliable external
acceptance signal, but not the same optimizer. The verifier supplies a binary
admission decision; ordinary LoRA SFT trains on complete accepted targets.
There is no policy-gradient update or reward optimization. A verifier is common
infrastructure with RLVR, not the novelty claim of this work.

\subsection{Verification and executable feedback}

Process supervision uses step-level human feedback to train reward models that
rank intermediate reasoning steps \citep{lightman2023verify}. In code
generation, CodeRL incorporates test and
critic feedback into training and regeneration \citep{le2022coderl}, while
LEVER learns to verify candidate programs from execution results
\citep{ni2023lever}. Self-Debugging uses execution signals and explanations to
revise generated programs \citep{chen2023selfdebug}. Our verifier is not learned
and the output is a structured plan rather than source code. This provides an
exact declared acceptance surface while restricting the claim to what that
surface expresses.

V-STaR trains a learned verifier from correct and incorrect self-generated
solutions and uses it to select among candidates at inference time
\citep{hosseini2024vstar}. Our selector is an external symbolic verifier used to
admit SFT targets rather than a learned verifier used for inference-time
ranking. The present negative selector replications do not test V-STaR's full
iterative method. They instead qualify the assumption that semantically cleaner
targets must yield a stable downstream capability advantage in a small-data
adapter regime.

\subsection{Language models and external planning verification}

PlanBench introduces an extensible benchmark for planning and reasoning about
change, including Blocksworld replanning
\citep{valmeekam2023planbench}. A separate critical investigation reports weak
autonomous plan generation and shows that sound validator feedback can support
back-prompting \citep{valmeekam2023critical}. Work on self-verification further
finds that intrinsic self-critique can degrade planning performance, while a
sound external verifier drives the measured gains
\citep{stechly2025selfverification}. Our external study adopts the PlanBench
repository's T6 replanning surface and pinned VAL
\citep{planbenchrepo,kclplanningval}, but asks a different question. Rather than
evaluate a
frozen model or use the verifier only for retry, we retain VAL-admitted runtime
plans as provenance-explicit supervision and evaluate the trained specialist
on fresh cases.

\subsection{Reflection and self-correction}

Agent methods such as Reflexion use verbal feedback and episodic memory to
improve later attempts \citep{shinn2023reflexion}. Other work documents limits
of intrinsic self-correction when no reliable external signal is present
\citep{huang2024selfcorrect,stechly2025selfverification}. Our inference-time
diagnostics are consistent with both observations: localized external residuals
can occasionally rescue a near miss, but they do not reliably create planning
fuel. The training studies ask whether independently verified histories can be
consolidated offline so the next attempt begins from a better specialized
policy.

\subsection{Position of this work}

The distinguishing combination is not LoRA, verification, or synthetic data in
isolation. It is the explicit separation of declared contract, model attempt,
verifier decision, accepted target, target producer, and training recipient,
followed by matched tests of admission, learning, iteration, cost, and stopping.
PlanBench and VAL supply the primary external task and verifier boundary. The
three workflow constructions supply controlled iteration and transfer
evidence. Independent enterprise authorship, institutional authority,
adjudication, and policy revision remain untested.

%% file: sections/12_conclusion.tex
\section{Conclusion}

Governance runtime records can be more than audit trails. When they preserve a
machine-checkable contract, model output, verifier decision, accepted target,
and model lineage, they can supply controlled supervision for a bounded
specialist.

The primary evidence comes from fresh external PlanBench replanning cases.
Qwen3-14B thinking and pinned VAL produce 24 admissible targets that train the
same checkpoint for non-thinking execution. The original adapter raises
accepted one-response plans from 1/80 to 57/80. In a prospective replication
that holds those targets, the model revision, recipe, and corpus fixed, all
eight new training seeds produce clear positive lifts under three inference
realizations each. Adapter results range from 45/80 to 70/80 against 1/80 for
every matched base report, with exact seed-level $p=0.0078125$. This is
teacher-free same-checkpoint consolidation without oracle labels. The original
paired interface-cure endpoint remains negative and must not be rewritten as a
separate cure claim.

The companion Phi result establishes a different pathway. Plans from a
separate reasoning teacher, selected by the same verifier, raise base Phi-4
from 2/80 to 51/80 verifier acceptance and from 35/80 to 80/80 schema validity.
This is stronger-teacher distillation, not self-training. The earlier synthetic
program supplies the teachability, cumulative-learning, replication, transfer,
and stopping evidence that bounds both external results.

The selector evidence draws a different boundary. At one matched seed, schema
selection, blinded model self-selection, and \valsys{} selection produce
55/160, 69/160, and 102/160 accepted plans. Across eight prospective seeds,
however, the \valsys{} versus model-self contrast changes sign and fails its
seed-level test ($p=0.3672$). The \valsys{} versus schema-only contrast has a
positive descriptive aggregate but also fails its reliability gate
($p=0.0703$). Independent verification is therefore load-bearing for target
admissibility and provenance, not for a demonstrated reliable capability
advantage over alternate selectors.

Within a machine-verifiable task contract, verifier-admitted model outputs can
therefore become supervision that robustly consolidates one occasionally
demonstrated capability into execution that is more consistent than base
non-thinking and far cheaper than fresh reasoning, while the verifier remains
the acceptance authority. The result does not establish general planning,
enterprise validity, unrestricted self-improvement, or model-size compression.
Those questions require new domains, new target harvests, institutional
evidence, and a direct specialist-size study.

%% file: sections/appendix_planbench.tex
\section{External PlanBench Study}
\label{app:planbench}

\subsection{Benchmark, verifier, and corpus}

The external surface uses PlanBench Blocksworld repository task T6, replanning
after an execution-time state change \citep{valmeekam2023planbench}. For
reproducibility, implementations are pinned as follows:
\begin{itemize}
    \item PlanBench repository:
    \par\texttt{1f4d600f4806bedbefebbbe44b168372aaf5060d}
    \citep{planbenchrepo}.
    \item VAL plan validator:
    \par\texttt{3c7a1f330bdab0ba28a4762bb45c3f06c27fb6d4}
    \citep{kclplanningval}.
    \item Fast Downward, used only during fresh-case construction:
    \par\texttt{824499f8f7b1d3f8c0260b2b8b0740ee815dcdca}.
\end{itemize}

The formal corpus contains 32 harvest cases and 80 evaluation cases. Each
partition is split evenly between 5 and 6 blocks. The builder excludes raw,
normalized-PDDL, and object-renaming-invariant structural overlap with 2,207
parseable published PlanBench problems, 150 prior experimental problem
files, and another formal partition. Harvest and evaluation have zero case and
structure overlap. Demonstrations are generated separately. Oracle replacement
plans are used transiently to establish case solvability but are not serialized
into prompts, target metadata, or retained training artifacts.

The frozen output interface is free text followed by a syntax-only,
token-conserving canonicalizer. It may wrap exact declared action and object
sequences in PDDL parentheses. It cannot add, remove, reorder, choose, or repair
actions. Every schema-valid candidate is submitted to VAL, which remains the
only semantic admission authority.

\subsection{Qualification history and protocol separation}

The positive formal study followed several frozen negative results. The first
non-thinking qualification found no VAL-accepted plan for either Qwen3-14B or
base Phi-4 on 24 cases within eight attempts. Qwen thinking later exposed
substantial occasional competence, but its strict final-plan interface failed
the original validity gate. A syntax-only diagnostic recovered additional
parseable attempts without qualifying a band.

A fresh fuel-band study then selected the 5--6 block band. On 24 confirmation
cases, Qwen thinking reached 6/24 on the first attempt and 17/24 within eight
attempts. Four responses exhausted the reasoning budget without a final plan,
and four more lacked a valid action sequence. The frozen result remained
\texttt{NO\_GO\_VALIDITY}; no adapter was authorized. A post-hoc audit found no
additional valid target among those eight failures.

A separately labeled smoke study trained only on the 17 VAL-admitted targets.
Its evaluation rows were case-disjoint from training, but the screen had
already influenced band selection. The smoke therefore supplied a directional
\texttt{SMOKE\_SIGNAL} and authorized a new protocol, not publication evidence.
The formal v1 protocol was then frozen, a new corpus was generated, and both
harvest and evaluation were rerun. This history is retained to prevent the
positive study from being read as a post-hoc waiver of earlier gates.

\subsection{Qwen formal protocol}

Qwen3-14B BF16 runs with native thinking enabled for up to eight independent
attempts per harvest case. Training is authorized only with at least 16
distinct VAL-admitted targets, at least eight pass-within-eight cases that fail
on the first attempt, both block-count strata, at least 98\% completed HTTP
requests, no target or partition contamination, at most 2.5 million completion
tokens, and at most six A100 hours. Invalid attempts cannot enter training, but
remain recorded against acquisition yield and cost.

The first VAL-accepted plan from each successful case is selected. Each target
record preserves case, attempt, seed, response hash, canonical plan hash, model
revision, serving regime, verifier commit, and verdict. The training recipient
is the same Qwen3-14B checkpoint with thinking disabled. The adapter uses BF16,
three fixed epochs, learning rate $10^{-4}$, 5\% warmup, cosine decay, LoRA rank
16, alpha 32, dropout 0.05, batch size one, and seed 2026113001. The final
checkpoint is selected after fixed exposure, not from evaluation performance.

The evaluation compares three one-response arms on the same 80 unopened cases:
base non-thinking, base thinking, and adapter non-thinking. The semantic
endpoint compares adapter with base non-thinking on VAL acceptance. It requires
at least 12 net gains, no more than five base-only regressions, exact two-sided
$p<0.05$, and success in both block strata. The co-primary interface endpoint
compares adapter with thinking on schema validity. It requires at least 95\%
adapter validity, zero endpoint exceptions, at least five paired gains, and
exact $p<0.05$.

\subsection{Qwen harvest and evaluation}

All Qwen harvest gates pass. Table~\ref{tab:qwen-external-harvest} reports the
acquisition result. The 24 admitted targets include 11 cases that succeed only
after the first attempt, cover both block counts, and have 24 distinct canonical
plan hashes. Training completes all 72 optimizer steps in 63.6 seconds.

\begin{table}[tbp]
\centering
\small
\begin{tabular}{lrr}
\toprule
Harvest measure & Observed & Frozen gate \\
\midrule
Distinct VAL-admitted targets & 24/32 & at least 16 \\
First-attempt successes & 13/32 & reported \\
Successes within eight attempts & 24/32 & reported \\
Beyond-first-attempt headroom & 11 & at least 8 \\
Completed HTTP requests & 118/118 & at least 98\% \\
Completion tokens & 419,201 & at most 2,500,000 \\
A100 time & 0.649 h & at most 6 h \\
\bottomrule
\end{tabular}
\caption{Formal Qwen target acquisition on the fresh harvest partition.}
\label{tab:qwen-external-harvest}
\end{table}

Table~\ref{tab:qwen-external-eval} gives the complete one-response evaluation.
Against base non-thinking, Self-24 records 56 adapter-only VAL successes, zero
base-only successes, one both-pass case, and 23 both-fail cases. The net gain is
56/80 with exact $p<10^{-10}$. The adapter-versus-thinking semantic table has
34 adapter-only, seven thinking-only, 23 both-pass, and 16 both-fail cases. Its
$p=0.0000253208$ is descriptive because superiority over thinking was not a
predeclared endpoint.

\begin{table}[tbp]
\centering
\small
\begin{tabular}{lrrrr}
\toprule
Qwen arm & VAL accepted & Schema-valid & Mean latency & Completion tokens \\
\midrule
Base, non-thinking & 1/80 & 71/80 & 0.624 s & 1,729 \\
Base, thinking & 30/80 & 77/80 & 65.964 s & 254,960 \\
Self-24, non-thinking & 57/80 & 80/80 & 1.186 s & 2,965 \\
\bottomrule
\end{tabular}
\caption{Fresh Qwen evaluation on the external 80-case partition. Token counts
are totals across each arm.}
\label{tab:qwen-external-eval}
\end{table}

The semantic endpoint passes in both strata. The interface endpoint does not:
Self-24 has three paired schema gains over thinking and no losses, but the
frozen gate requires five gains and exact $p<0.05$; observed $p=0.25$. The
combined reading is \texttt{SEMANTIC\_ONLY}. The adapter's absolute deployment
interface is nevertheless perfect on the measured set, with 80/80 schema-valid
responses and zero endpoint errors.

\subsection{Matched verifier-selection ablation}

The ablation was frozen to isolate target admission from candidate generation.
It reuses the 24 successful Qwen harvest cases and all 52 schema-valid
candidates from those cases. Thirteen cases have one eligible candidate. For
the 11 contested cases, Qwen thinking sees only opaque, shuffled candidate IDs
and must select one without rewriting, repairing, or combining plans. Its
packet contains no VAL outcome, attempt index, generation seed, response hash,
candidate provenance, or stop signal. A post-run leak scan is clean. The
schema-only arm takes the first candidate in the same frozen opaque ordering.

Every arm holds fixed the 24 case IDs, candidate pool, prompt surface, target
count, official Qwen3-14B BF16 base, LoRA rank 16, alpha 32, dropout 0.05,
learning rate $10^{-4}$, three warmup steps, three epochs, 72 optimizer steps,
and seed 2026120301. Each trains on one A100 80 GB PCIe. Final checkpoints are
selected by fixed exposure rather than evaluation. Table
\ref{tab:selection-targets} reports the manipulation check and training
outcomes.

\begin{table}[tbp]
\centering
\small
\begin{tabular}{lrrr}
\toprule
Selector & VAL-valid targets & Train time & Final loss \\
\midrule
VAL & 24/24 & 64.1 s & 0.3545 \\
Qwen self-selection & 20/24 & 64.2 s & 0.3905 \\
Schema-only frozen order & 15/24 & 64.0 s & 0.5077 \\
\bottomrule
\end{tabular}
\caption{Matched target sets and training outcomes. Only target selection
changes across arms.}
\label{tab:selection-targets}
\end{table}

Evaluation uses 160 newly generated and structurally excluded replanning
cases, evenly split between 5 and 6 blocks. No replacement plan is serialized
into model-visible material. All 640 calls complete without endpoint errors.
Table~\ref{tab:selection-eval} gives the full arm outcomes.

\begin{table}[tbp]
\centering
\small
\begin{tabular}{lrrr}
\toprule
Evaluation arm & VAL accepted & Schema-valid & Mean latency \\
\midrule
Base, non-thinking & 1/160 & 129/160 & 1.395 s \\
Schema-selected adapter & 55/160 & 152/160 & 1.410 s \\
Model-self-selected adapter & 69/160 & 159/160 & 1.385 s \\
VAL-selected adapter & 102/160 & 160/160 & 1.202 s \\
\bottomrule
\end{tabular}
\caption{Fresh outcomes for the matched target-selection ablation.}
\label{tab:selection-eval}
\end{table}

The frozen primary comparison is VAL selection versus model self-selection. In
all 160 cases it has 41 VAL-only successes, eight self-only successes, paired
net $+33$, and exact two-sided $p=1.9647\times10^{-6}$. The 5-block stratum is
54/80 versus 44/80, with paired net $+10$ and $p=0.0308837891$. The 6-block
stratum is 48/80 versus 25/80, with paired net $+23$ and
$p=3.39532\times10^{-5}$. Every frozen gate passes. Secondary comparisons
also favor model self-selection over schema-only selection by paired net $+14$
with $p=0.03848$, and VAL over schema-only selection by paired net $+47$ with
$p=4\times10^{-10}$.

The frozen reading is \texttt{VERIFIER\_SELECTION\_ADVANTAGE}. It permits a
causal statement about target selection for this seed and matched comparison.
It does not make the verifier the sole cause of the full training effect,
establish an exact purity-response law, or generalize to other training seeds
or open-world correctness. The experiment uses one model family, one
harvestable 24-case pool, one adapter seed, one domain, and one difficulty
band. The prospective results below show that its beneficial direction is not
reliable across training seeds.

Before any model call, the clean runtime received the protocol-pinned
\texttt{tarski==0.7.0} and \texttt{pddl==0.2.0} dependencies. A frozen corpus
wrapper directory precondition was handled by running its underlying frozen
builder configuration directly and reproducing manifest validation. These
plumbing amendments changed no seed, exclusion, case count, model call, target,
training control, or gate.

\subsection{Prospective Self-24 training-seed replication}

The prospective primary replication reuses the exact 24 Qwen thinking targets
admitted in the parent study. It holds the Qwen3-14B revision, rank-16 LoRA
recipe, 72 optimizer steps, 80-case evaluation corpus, prompts, decoding
surface, and pinned \valsys{} implementation fixed. Seeds 2026113002 through
2026113009 are new training runs. Each adapter and a matched regenerated base
are evaluated under three frozen inference-seed bases. The archived parent
adapter is rerun only as context and excluded from the prospective test.

\begin{table}[tbp]
\centering
\small
\begin{tabular}{lrrrr}
\toprule
Training seed & Adapter reports & Base reports & Mean net & Reading \\
\midrule
2026113002 & 59, 61, 59 & 1, 1, 1 & $+58.67$ & clear lift \\
2026113003 & 57, 59, 57 & 1, 1, 1 & $+56.67$ & clear lift \\
2026113004 & 45, 47, 45 & 1, 1, 1 & $+44.67$ & clear lift \\
2026113005 & 51, 52, 51 & 1, 1, 1 & $+50.33$ & clear lift \\
2026113006 & 59, 58, 57 & 1, 1, 1 & $+57.00$ & clear lift \\
2026113007 & 65, 65, 66 & 1, 1, 1 & $+64.33$ & clear lift \\
2026113008 & 58, 59, 58 & 1, 1, 1 & $+57.33$ & clear lift \\
2026113009 & 66, 68, 70 & 1, 1, 1 & $+67.00$ & clear lift \\
\bottomrule
\end{tabular}
\caption{Accepted plans out of 80 for the eight prospective Self-24 training
seeds. Each report column contains the three frozen inference realizations.}
\label{tab:self24-eight-seed}
\end{table}

All eight seeds are clear lifts, none is null or reverse, and the exact
two-sided sign-flip test gives $p=0.0078125$. Across the 24 reports, adapters
produce 1392/1920 accepted plans and bases produce 24/1920. These pooled totals
are descriptive. The seed-level effects are the inferential observations.
Every report improves both block-count strata, all 3840 requests are
transport-valid, and no report contains a base-only semantic regression.
Adapter schema validity is 1910/1920, compared with 1662/1920 for base.

The result is \texttt{PRIMARY\_SELF\_TRAINING\_REPLICATES}. It establishes
training-seed robustness for one fixed intervention. It does not establish
robustness to new targets, evaluation cases, recipes, model revisions, or
domains.

\subsection{Prospective selector robustness studies}

The selector extensions reuse the same \valsys{}-selected and model-self
partitions as the parent ablation and prospectively train seeds 2026120305
through 2026120312. Each adapter is evaluated under three inference
realizations on the same 160-case corpus. A second study trains schema-only
adapters under the same seeds and compares them with the archived \valsys{}
arms. Table~\ref{tab:selector-eight-seed} reports common-seed totals. Each cell
is out of 480 repeated outcomes, three realizations over 160 cases.

\begin{table}[tbp]
\centering
\small
\begin{tabular}{lrrrrr}
\toprule
Seed & Schema & Model self & \valsys{} & \valsys{}$-$self & \valsys{}$-$schema \\
\midrule
2026120305 & 102 & 123 & 190 & $+67$ & $+88$ \\
2026120306 & 164 & 120 & 141 & $+21$ & $-23$ \\
2026120307 & 129 & 248 & 288 & $+40$ & $+159$ \\
2026120308 & 147 & 151 & 118 & $-33$ & $-29$ \\
2026120309 & 110 & 197 & 226 & $+29$ & $+116$ \\
2026120310 & 146 & 215 & 149 & $-66$ & $+3$ \\
2026120311 & 148 & 180 & 187 & $+7$ & $+39$ \\
2026120312 & 102 & 139 & 198 & $+59$ & $+96$ \\
\midrule
Total & 1048 & 1373 & 1497 & $+124$ & $+449$ \\
\bottomrule
\end{tabular}
\caption{Prospective common-seed selector results. Totals are descriptive;
exact tests use the eight training-seed effects rather than pooled cases.}
\label{tab:selector-eight-seed}
\end{table}

For \valsys{} versus model self-selection, two seeds show clear \valsys{}
effects, one shows a clear model-self effect, and five are null. The exact
seed-level sign-flip value is $p=0.3671875$.

\noindent Frozen reading:
\texttt{NO\_REPLICATED\_SELECTOR\_ADVANTAGE}. For \valsys{} versus
schema-only, four seeds are clear \valsys{} effects, four are null, and none is
a clear schema effect. The primary rule requires at least six clear \valsys{}
seeds and $p<0.05$; observed $p=0.0703125$, giving
\texttt{NO\_REPLICATED\_SEMANTIC\_SELECTION\_ADVANTAGE}.

Each frozen selector comparison contains 7,680 transport-valid and
endpoint-clean request outcomes; the schema comparison reuses the archived
\valsys{} reports rather than regenerating that arm. Semantic scoring is
performed after generation from hash-sealed response bytes with the pinned
\valsys{} executable and shared library. A preliminary three-seed evaluation is excluded because
the verifier library was missing and the evaluator converted loader failures
into semantic rejection. The exact response bytes were preserved and rescored
after unexpected process status became fatal. This incident changed no model
response and has no standing as a performance result.

The original single-seed verdict remains a valid conditional result. The
prospective evidence changes its general interpretation: semantic selection
does not have a replicated downstream capability advantage over either
alternative under this fixed low-data recipe. Only the \valsys{} partition,
however, carries 24/24 independent semantic admission receipts.

\subsection{Phi protocol, audit, and evaluation}

The companion Phi protocol reuses the same corpus, prompts, verifier, fixed
training recipe, and gates. It was frozen before the formal Qwen response
partitions were opened. Phi-4-reasoning is the target producer and base Phi-4
is the training recipient. This is governance-selected stronger-teacher
distillation.

The pinned teacher template requests a reasoning section followed by a final
solution. A frozen delimiter-only extractor returns only text after the last
closing reasoning delimiter. Responses that never close the reasoning section
are rejected, and action-like text embedded in reasoning is never recovered.
The extractor cannot consult VAL, action legality, or alternative parses.

Phi harvest admits 24/32 distinct targets, with 10 first-attempt successes and
24 successes within eight attempts. All 256 requests complete, acquisition
uses 1,725,255 completion tokens and 2.791 A100 hours, and accepted targets
cover both strata. Training completes 72 optimizer steps in 56.8 seconds.

The first training invocation stopped before model load because the remote
bundle omitted a local trainer dependency. Harvest, extraction, targets, and
the unopened evaluation corpus were already sealed. A machine-readable resume
amendment added the missing hashed dependency, reverified the exact training
file and partition isolation, and changed no model revision, case, target,
prompt, seed, hyperparameter, gate, or reading.

\begin{table}[tbp]
\centering
\small
\begin{tabular}{lrrrr}
\toprule
Phi arm & VAL accepted & Schema-valid & Mean latency & Completion tokens \\
\midrule
Base Phi-4 & 2/80 & 35/80 & 1.027 s & 3,362 \\
Phi-4-reasoning & 30/80 & 41/80 & 150.283 s & 544,408 \\
Distilled Phi-4 & 51/80 & 80/80 & 0.981 s & 2,531 \\
\bottomrule
\end{tabular}
\caption{Fresh Phi evaluation on the same external 80-case partition. Token
counts are totals across each arm.}
\label{tab:phi-external-eval}
\end{table}

Against base Phi-4, the adapter records 49 adapter-only VAL successes, zero
base-only successes, two both-pass cases, and 29 both-fail cases. The net gain
is 49/80 with exact $p<10^{-10}$. Schema validity improves by 39 paired cases
over the reasoning teacher, with zero losses and exact $p<10^{-10}$. Both
semantic strata pass their frozen gates. The reading is
\texttt{FULL\_BAND\_DISTILLATION}.

The distilled student's 51/80 versus the teacher's 30/80 has 30 student-only,
nine teacher-only, 21 both-pass, and 20 both-fail cases. The exact paired value
is $p=0.0010650196$, but the comparison remains a compression diagnostic rather
than a preregistered teacher-superiority test.

%% file: sections/appendix_synthetic_program.tex
\section{Synthetic Development Program}
\label{app:synthetic-program}

The synthetic studies supplied mechanism, replication, and boundary evidence
before the external experiment. They are supporting evidence rather than the
paper's primary result. This appendix therefore organizes them by claim instead
of reproducing the chronological experiment log. Generator A, Generator B, and
Generator C use different self-authored construction code, vocabularies, and
mutation logic, but share the public plan schema, deterministic verifier, base
model families, and researcher. Results are never pooled across constructions.

\subsection{Study map and controls}

Table~\ref{tab:synthetic-study-map} distinguishes the target source and purpose
of each stage. In the oracle precursor, procedural schedules supply the targets
and the verifier confirms them. In every self-sourced stage, the model receives
only the malformed plan, public contract, feedback, and output schema. A target
is admitted only when a sampled model response is parseable and passes the
frozen verifier. No failed row is filled with a procedural answer. Cumulative
adapters are reinitialized from the original base and trained on the cumulative
admitted set rather than continued from the preceding adapter.

\begin{table}[tbp]
\centering
\footnotesize
\begin{tabular}{@{}p{0.15\linewidth}p{0.12\linewidth}p{0.21\linewidth}p{0.17\linewidth}p{0.22\linewidth}@{}}
\toprule
Stage & Model & Training-target source & Fresh evaluation & Evidentiary role \\
\midrule
Oracle precursor & Phi-4, Qwen3-14B & Procedural schedule, verifier-confirmed & 144 rows per model & Teachability and signal attribution \\
Generator A & Qwen3-14B & Same-model accepted samples & 48-row one-response and 32-row search pools & Bounded loop, cumulative learning, stopping \\
Generator B & Qwen3-14B & Same-model accepted samples & 80-row one-response and 32-row search pools & Native construction replication and transfer boundary \\
Generator C & Qwen3-14B & Same-model accepted samples & 128-row one-response and 32-row search pools & Construction-separated first-round replication \\
Generator B & Phi-4 & Same-model accepted samples & 80-row one-response and 32-row search pools & Cross-model method replication \\
\bottomrule
\end{tabular}
\caption{Synthetic evidence map. All evaluation partitions are case-disjoint
from their training targets. Generator separation is by implementation, not by
independent authorship.}
\label{tab:synthetic-study-map}
\end{table}

Unless stated otherwise, the main endpoint is one-response verifier acceptance
on an unopened partition. Search studies additionally report success within a
fixed 32-attempt budget, denoted pass@32. Paired changes use exact two-sided
McNemar tests. Frozen gates also check transport validity, minimum effect size,
and family-level regressions. These tests describe the frozen samples and are
not case-clustered population estimates. The oracle Phi and Qwen studies, and
all later Qwen and Phi self-sourcing studies, use BF16 on one A100 80GB GPU.
One training seed is used for each reported adapter.

\subsection{Oracle teachability and learning-signal attribution}

The precursor holds the corpus, verifier, target plans, training recipe, and
one-response evaluation fixed across two model families. Phi-4 improves from
33/144 to 137/144 accepted repairs, with 104 adapter-only passes and no
base-only passes. Qwen3-14B improves from 31/144 to 139/144, with 108
adapter-only passes and no base-only passes. On the 48-row compound subset,
Phi moves from 6/48 to 46/48 and Qwen from 5/48 to 45/48. Every compound family
appears in training, so this establishes in-distribution teachability rather
than held-out-family composition.

The target audit limits that result. The generator first creates one canonical
schedule for each case, constructs a compatible public contract, and mutates a
copy of that schedule. In training, 192 compound rows across 48 cases reuse 48
unique case-level targets. The verifier confirms each target but neither
searches for nor generates it. The correct classification is therefore
oracle-target SFT with deterministic verification, not oracle-free learning.

The matched Qwen feedback cross in Table~\ref{tab:synthetic-ablation} asks how
much of the lift comes from complete accepted targets versus localized trace
detail. Under the same localized inference interface, localized-trace training
scores 139/144 and verdict-only training scores 133/144. The 4.2 percentage
point difference lies within the frozen five-point boundary, and verdict-only
training remains above the 90\% sufficiency threshold. The paired table contains
132 both-pass rows, four both-fail rows, seven localized-only rows, and one
verdict-only-only row. Complete accepted targets therefore carry most of the
measured teaching signal in distribution, while localized detail is additive.

\begin{table}[tbp]
\centering
\small
\begin{tabular}{llrrr}
\toprule
Training state & Inference feedback & Pass & Transport & Compound \\
\midrule
Base & Localized & 31/144 & 128/144 & 5/48 \\
Localized-trained & Localized & 139/144 & 144/144 & 45/48 \\
Base & Verdict-only & 35/144 & 135/144 & 9/48 \\
Localized-trained & Verdict-only & 115/144 & 144/144 & 33/48 \\
Verdict-only-trained & Localized & 133/144 & 144/144 & 41/48 \\
Verdict-only-trained & Verdict-only & 136/144 & 144/144 & 44/48 \\
\bottomrule
\end{tabular}
\caption{Qwen training and inference feedback cross. The primary causal
comparison fixes localized inference and changes only the training trace.}
\label{tab:synthetic-ablation}
\end{table}

Localization can still matter at inference. In a separate Phi diagnostic,
exact residuals produced 3/15 verified completions while a generic retry
produced none. All three were in one co-location family. This narrow diagnostic
does not change the training attribution above.

\subsection{Verifier-selected learning, iteration, and stopping}

The first target-hidden Qwen qualification tests whether a model can supply
its own labels. At $N=32$, fixed localized feedback yields 16 accepted targets
from 32 cases. Ten pass on the first attempt and six appear only after earlier
rejections. Table~\ref{tab:synthetic-search-attribution} shows the matched
search controls. Localization adds four net harvests over generic resampling
but is not significant in this small paired sample ($p=0.34375$). Adaptive
residual updates reduce calls without expanding coverage. In all three arms,
the verifier is load-bearing as the admission rule, not as a target generator.

\begin{table}[tbp]
\centering
\small
\begin{tabular}{lrrrrr}
\toprule
Search condition & Accepted & Pass@1 & Later & Calls & Transport \\
\midrule
Fixed generic & 12/32 & 8/32 & 4 & 678 & 678/678 \\
Fixed localized & 16/32 & 10/32 & 6 & 591 & 590/591 \\
Adaptive localized & 16/32 & 11/32 & 5 & 539 & 539/539 \\
\bottomrule
\end{tabular}
\caption{Target-hidden search attribution. ``Later'' counts targets first
accepted after at least one verifier rejection.}
\label{tab:synthetic-search-attribution}
\end{table}

Training on the 16 Qwen-generated targets raises fresh Generator A
one-response repair from 10/48 to 21/48. The paired transition contains 14
gains and three losses ($p=0.012726$). A coverage-matched adapter trained on 16
procedural targets reaches 23/48 and is not distinguishable from Self-16 in
this small sample ($p=0.726562$). This closes the first bounded loop from model
sampling, through verifier admission and LoRA training, to fresh same-model
improvement.

Table~\ref{tab:synthetic-cumulative} condenses the cumulative results and their
stopping points. On Generator A, 14 additional Self-16 targets produce the
30-target Default30 adapter. It improves over Self-16 on two separate fresh
partitions and expands fixed-budget reachability from 13/32 to 22/32, with nine
paired gains and no losses ($p=0.003906$). A 52-target third adapter reaches
35/48 versus Default30 at 33/48, but its three gains and one loss give only a
net two ($p=0.625$), below the frozen four-row continuation threshold. Round 4
is therefore blocked.

\begin{table}[tbp]
\centering
\scriptsize
\begin{tabular}{@{}p{0.17\linewidth}p{0.22\linewidth}p{0.23\linewidth}p{0.30\linewidth}@{}}
\toprule
Construction and model & Earlier arm & Later arm & Paired reading \\
\midrule
A, Qwen initial & Base 10/48 & Self-16 21/48 & 14 gains, 3 losses, $p=0.012726$ \\
A, Qwen Round 2 & Self-16 25/48 & Default30 35/48 & Net $+10$, $p=0.021271$ \\
A, Qwen Round 3 & Default30 33/48 & Round3 35/48 & 3 gains, 1 loss, $p=0.625$; stop \\
B, Qwen Round 1 & Base 12/80 & B-Round1 44/80 & Net $+32$, $p=4.07\times10^{-9}$ \\
B, Qwen Round 2 & B-Round1 44/80 & B-Round2 62/80 & Net $+18$, $p=4.01\times10^{-5}$ \\
C, Qwen final & Base 57/128 & C-Round2 126/128 & 69 gains, 0 losses, $p=3.388\times10^{-21}$ \\
B, Phi round & Base 10/80 & Self-sourced LoRA 77/80 & 67 gains, 0 losses, $p=1.3553\times10^{-20}$ \\
\bottomrule
\end{tabular}
\caption{Fresh one-response transitions from verifier-selected self-training.
The Generator A rows use separate case-disjoint partitions. Generator C
Round 1 reaches 128/128; its final Round 2 arm reaches 126/128 and does not
improve on the saturated first-round measurement.}
\label{tab:synthetic-cumulative}
\end{table}

Generator B supplies a second cumulative construction. Qwen harvests 21/64
Round-1 targets, then 59 cumulative targets after the second harvest. Fresh
one-response repair rises from 12/80 to 44/80 and then 62/80. The four family
trajectories are 4/20 to 8/20 to 8/20 for duplicate plus colocation, 3/20 to
18/20 to 20/20 for omission plus unauthorized execution, 5/20 to 18/20 to
19/20 for precedence plus control, and 0/20 to 0/20 to 15/20 for timing plus
separation. The second round therefore adds a capability family not improved
by the first round.

\begin{figure}[tbp]
\centering
\includegraphics[width=\linewidth]{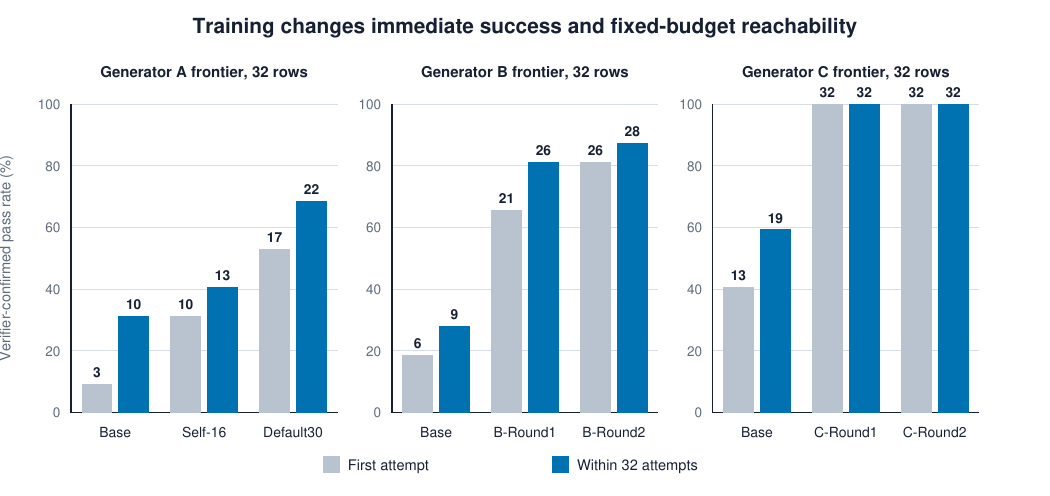}
\caption{First-attempt and 32-attempt success on matched pools. Panels use
different case sets and are not pooled. Training expands fixed-budget
reachability through Default30 on Generator A and Round 1 on Generator B;
Generator C saturates the frozen frontier after Round 1.}
\label{fig:search-frontiers}
\end{figure}

Figure~\ref{fig:search-frontiers} separates improved immediacy from expanded
fixed-budget reachability. On Generator A, base, Self-16, and Default30 reach
10/32, 13/32, and 22/32 within 32 attempts. On Generator B, base, Round 1, and
Round 2 reach 9/32, 26/32, and 28/32. The first B round adds a paired net 17
($p=0.0000153$); the second adds only net two ($p=0.625$), below the frozen
frontier continuation threshold even though its separate one-response gain is
material. Generator C rises from 19/32 to 32/32 after Round 1, so that frontier
has no remaining room to measure Round-2 expansion.

Two null branches constrain the iteration account. On an earlier 32-row
Generator A pool, base and Self-16 both reach 14/32 at pass@32, so the frozen
gate blocks training. Self-16 nevertheless raises first-attempt success from 3
to 10 and reduces calls from 697 to 622. On their 14 shared-unresolved rows,
mean unique plans fall from 13.71 to 5.14, duplicate rate rises from 57.1\% to
83.9\%, and residual entropy falls from 0.438 to 0.214. A later temperature
study increases pass@32 from 14/24 at $T=0.7$ to 18/24 at $T=1.3$ but misses
the frozen diversity-transfer gate and is not significant ($p=0.125$).
Therefore neither the plateau nor higher temperature supports an unrestricted
iteration rule.

Generator C provides a construction-separated replication at a higher base
rate. Round 1 harvests 52/64 targets. On the hidden evaluation, base, Round 1,
and Round 2 score 57/128, 128/128, and 126/128. The preregistered Round-2 versus
base gain is 53.9 points, with a row-bootstrap interval of [45.3, 62.5]. Round
2 loses two rows relative to Round 1 ($p=0.5$), and both adapters reach 32/32 on
the frontier. The frozen measurements saturate; the research program does not.
A seed-handling defect in the initial serving run prompted a transport-only
amendment before scientific calls. It changed no prompt, admission rule,
partition, gate, or outcome, and all final responses are schema-valid.

The Phi Generator B run tests the same chain in another model family. Phi
harvests 17/64 targets, including ten found only after earlier rejections and
targets from all four families. The LoRA moves fresh one-response repair from
10/80 to 77/80 and the fixed-budget frontier from 9/32 to 26/32, a paired net
17 ($p=7.629\times10^{-5}$). A comparator amendment moved a transport predicate
to its frozen protocol location without changing outputs, counts, thresholds,
or readings. Because the Qwen outcomes were already known, this is
fixed-benchmark cross-model replication, not a controlled model ranking.

\subsection{Transfer and explanatory boundaries}

The zero-shot transfer test applies Generator A adapters to 160 unopened
Generator B rows. The source anchor first reproduces the A effect, from 11/48
for base to 33/48 for Default30. On Generator B, base, Self-16, Default30, and
Round3 score 28/160, 29/160, 47/160, and 55/160. Default30 is the frozen primary
treatment. It produces 24 gains and five losses relative to base, a paired net
19 ($p=0.000546$), but crosses the family-regression guard: precedence plus
control falls from 10/40 to 6/40. Omission plus unauthorized repair rises from
9/40 to 22/40, duplicate plus colocation rises from 9/40 to 19/40, and every
arm remains 0/40 on timing plus separation. The earned primary reading is
selective transfer with a family regression.

\begin{figure}[!htbp]
\centering
\includegraphics[width=0.82\linewidth]{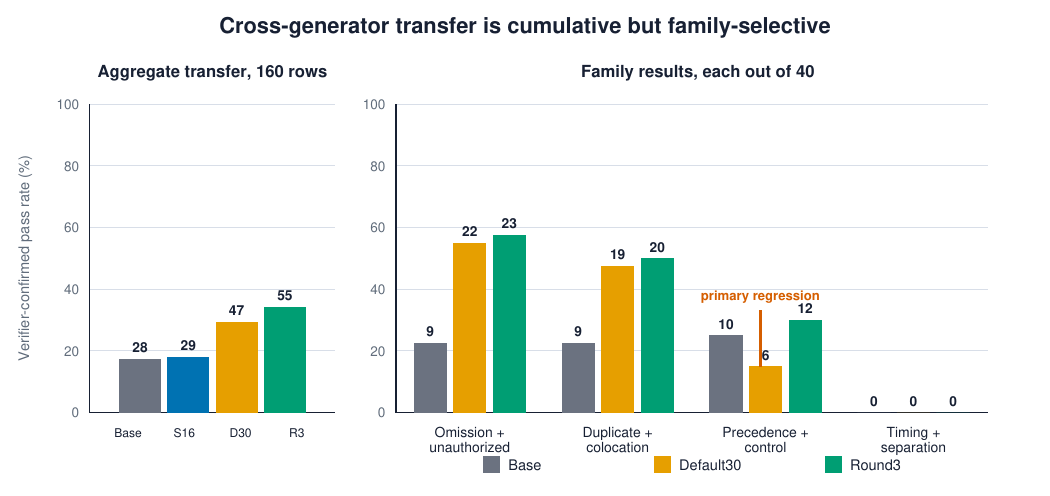}
\caption{Zero-shot transfer from Generator A adapters to Generator B.
Default30 passes the aggregate gate but crosses the frozen family-regression
guard. Round3 is a predeclared secondary trajectory arm and cannot replace the
primary treatment after results are opened.}
\label{fig:transfer-selectivity}
\end{figure}

Round3 reaches 55/160, with family nets of $+14$, $+11$, $+2$, and zero versus
base ($p=0.00000743$); its net eight over Default30 is not significant
($p=0.1153$). Native B training resolves failures that transfer does not,
including timing plus separation, so transfer is not a substitute for target
support from the destination distribution.

\enlargethispage{2\baselineskip}
A post-hoc local repair-graph audit finds no scalar separator. Overall, the
program supports large first-round gains across three self-authored
constructions and two model families, cumulative gains in A and B, and expanded
fixed-budget reachability. Its measured stops, selective transfer, and
saturated evaluations rule out unrestricted improvement or general enterprise
validity.

%% file: references.bib
@inproceedings{hu2022lora,
  author    = {Edward J. Hu and Yelong Shen and Phillip Wallis and Zeyuan Allen-Zhu and Yuanzhi Li and Shean Wang and Lu Wang and Weizhu Chen},
  title     = {{LoRA}: Low-Rank Adaptation of Large Language Models},
  booktitle = {International Conference on Learning Representations},
  year      = {2022},
  url       = {https://arxiv.org/abs/2106.09685}
}

@misc{dettmers2023qlora,
  author        = {Tim Dettmers and Artidoro Pagnoni and Ari Holtzman and Luke Zettlemoyer},
  title         = {{QLoRA}: Efficient Finetuning of Quantized {LLM}s},
  year          = {2023},
  eprint        = {2305.14314},
  archiveprefix = {arXiv},
  primaryclass  = {cs.LG},
  url           = {https://arxiv.org/abs/2305.14314}
}

@misc{zelikman2022star,
  author        = {Eric Zelikman and Yuhuai Wu and Jesse Mu and Noah D. Goodman},
  title         = {{STaR}: Bootstrapping Reasoning With Reasoning},
  year          = {2022},
  eprint        = {2203.14465},
  archiveprefix = {arXiv},
  primaryclass  = {cs.LG},
  url           = {https://arxiv.org/abs/2203.14465}
}

@inproceedings{wang2023selfinstruct,
  author    = {Yizhong Wang and Yeganeh Kordi and Swaroop Mishra and Alisa Liu and Noah A. Smith and Daniel Khashabi and Hannaneh Hajishirzi},
  title     = {Self-Instruct: Aligning Language Models with Self-Generated Instructions},
  booktitle = {Proceedings of the 61st Annual Meeting of the Association for Computational Linguistics},
  year      = {2023},
  url       = {https://arxiv.org/abs/2212.10560}
}

@misc{yuan2023scaling,
  author        = {Zheng Yuan and Hongyi Yuan and Chengpeng Li and Guanting Dong and Keming Lu and Chuanqi Tan and Chang Zhou and Jingren Zhou},
  title         = {Scaling Relationship on Learning Mathematical Reasoning with Large Language Models},
  year          = {2023},
  eprint        = {2308.01825},
  archiveprefix = {arXiv},
  primaryclass  = {cs.CL},
  url           = {https://arxiv.org/abs/2308.01825}
}

@misc{gulcehre2023rest,
  author        = {Caglar Gulcehre and Tom Le Paine and Srivatsan Srinivasan and Ksenia Konyushkova and Lotte Weerts and Abhishek Sharma and Aditya Siddhant and Alex Ahern and Miaosen Wang and Chenjie Gu and Wolfgang Macherey and Arnaud Doucet and Orhan Firat and Nando de Freitas},
  title         = {Reinforced Self-Training ({ReST}) for Language Modeling},
  year          = {2023},
  eprint        = {2308.08998},
  archiveprefix = {arXiv},
  primaryclass  = {cs.CL},
  url           = {https://arxiv.org/abs/2308.08998}
}

@misc{shao2024deepseekmath,
  author        = {Zhihong Shao and Peiyi Wang and Qihao Zhu and Runxin Xu and Junxiao Song and Xiao Bi and Haowei Zhang and Mingchuan Zhang and Y. K. Li and Y. Wu and Daya Guo},
  title         = {{DeepSeekMath}: Pushing the Limits of Mathematical Reasoning in Open Language Models},
  year          = {2024},
  eprint        = {2402.03300},
  archiveprefix = {arXiv},
  primaryclass  = {cs.CL},
  url           = {https://arxiv.org/abs/2402.03300}
}

@article{deepseekai2025r1,
  author  = {{DeepSeek-AI} and others},
  title   = {{DeepSeek-R1}: Incentivizing Reasoning Capability in {LLM}s via Reinforcement Learning},
  journal = {Nature},
  volume  = {645},
  pages   = {633--638},
  year    = {2025},
  doi     = {10.1038/s41586-025-09422-z},
  url     = {https://arxiv.org/abs/2501.12948}
}

@inproceedings{chen2024spin,
  author    = {Zixiang Chen and Yihe Deng and Huizhuo Yuan and Kaixuan Ji and Quanquan Gu},
  title     = {Self-Play Fine-Tuning Converts Weak Language Models to Strong Language Models},
  booktitle = {International Conference on Machine Learning},
  year      = {2024},
  url       = {https://arxiv.org/abs/2401.01335}
}

@misc{lightman2023verify,
  author        = {Hunter Lightman and Vineet Kosaraju and Yura Burda and Harri Edwards and Bowen Baker and Teddy Lee and Jan Leike and John Schulman and Ilya Sutskever and Karl Cobbe},
  title         = {Let's Verify Step by Step},
  year          = {2023},
  eprint        = {2305.20050},
  archiveprefix = {arXiv},
  primaryclass  = {cs.LG},
  url           = {https://arxiv.org/abs/2305.20050}
}

@misc{hosseini2024vstar,
  author        = {Arian Hosseini and Xingdi Yuan and Nikolay Malkin and Aaron Courville and Alessandro Sordoni and Rishabh Agarwal},
  title         = {{V-STaR}: Training Verifiers for Self-Taught Reasoners},
  year          = {2024},
  eprint        = {2402.06457},
  archiveprefix = {arXiv},
  primaryclass  = {cs.LG},
  url           = {https://arxiv.org/abs/2402.06457}
}

@inproceedings{ni2023lever,
  author    = {Ansong Ni and Srini Iyer and Dragomir Radev and Ves Stoyanov and Wen-tau Yih and Sida I. Wang and Xi Victoria Lin},
  title     = {{LEVER}: Learning to Verify Language-to-Code Generation with Execution},
  booktitle = {International Conference on Machine Learning},
  year      = {2023},
  url       = {https://arxiv.org/abs/2302.08468}
}

@inproceedings{le2022coderl,
  author    = {Hung Le and Yue Wang and Akhilesh Deepak Gotmare and Silvio Savarese and Steven C. H. Hoi},
  title     = {{CodeRL}: Mastering Code Generation through Pretrained Models and Deep Reinforcement Learning},
  booktitle = {Advances in Neural Information Processing Systems},
  year      = {2022},
  url       = {https://arxiv.org/abs/2207.01780}
}

@misc{chen2023selfdebug,
  author        = {Xinyun Chen and Maxwell Lin and Nathanael Scharli and Denny Zhou},
  title         = {Teaching Large Language Models to Self-Debug},
  year          = {2023},
  eprint        = {2304.05128},
  archiveprefix = {arXiv},
  primaryclass  = {cs.CL},
  url           = {https://arxiv.org/abs/2304.05128}
}

@inproceedings{shinn2023reflexion,
  author    = {Noah Shinn and Federico Cassano and Edward Berman and Ashwin Gopinath and Karthik Narasimhan and Shunyu Yao},
  title     = {Reflexion: Language Agents with Verbal Reinforcement Learning},
  booktitle = {Advances in Neural Information Processing Systems},
  volume    = {36},
  year      = {2023},
  url       = {https://arxiv.org/abs/2303.11366}
}

@inproceedings{huang2024selfcorrect,
  author    = {Jie Huang and Xinyun Chen and Swaroop Mishra and Huaixiu Steven Zheng and Adams Wei Yu and Xinying Song and Denny Zhou},
  title     = {Large Language Models Cannot Self-Correct Reasoning Yet},
  booktitle = {International Conference on Learning Representations},
  year      = {2024},
  url       = {https://openreview.net/forum?id=IkmD3fKBPQ}
}

@inproceedings{stechly2025selfverification,
  author    = {Kaya Stechly and Karthik Valmeekam and Subbarao Kambhampati},
  title     = {On the Self-Verification Limitations of Large Language Models on Reasoning and Planning Tasks},
  booktitle = {International Conference on Learning Representations},
  year      = {2025},
  url       = {https://proceedings.iclr.cc/paper_files/paper/2025/file/f3c5e56274140e0420baa3916c529210-Paper-Conference.pdf}
}

@inproceedings{valmeekam2023planbench,
  author    = {Karthik Valmeekam and Matthew Marquez and Alberto Olmo and Sarath Sreedharan and Subbarao Kambhampati},
  title     = {{PlanBench}: An Extensible Benchmark for Evaluating Large Language Models on Planning and Reasoning about Change},
  booktitle = {Advances in Neural Information Processing Systems},
  volume    = {36},
  year      = {2023},
  note      = {Datasets and Benchmarks Track},
  url       = {https://arxiv.org/abs/2206.10498}
}

@inproceedings{valmeekam2023critical,
  author    = {Karthik Valmeekam and Matthew Marquez and Sarath Sreedharan and Subbarao Kambhampati},
  title     = {On the Planning Abilities of Large Language Models: A Critical Investigation},
  booktitle = {Advances in Neural Information Processing Systems},
  volume    = {36},
  year      = {2023},
  url       = {https://arxiv.org/abs/2305.15771}
}

@misc{planbenchrepo,
  author       = {Karthik Valmeekam and Matthew Marquez and Alberto Olmo and Sarath Sreedharan and Subbarao Kambhampati},
  title        = {{PlanBench} Software and Data Repository},
  year         = {2026},
  howpublished = {Software repository},
  note         = {Pinned in this study at commit 1f4d600f4806bedbefebbbe44b168372aaf5060d},
  url          = {https://github.com/karthikv792/LLMs-Planning}
}

@misc{kclplanningval,
  author       = {{KCL Planning Group}},
  title        = {{VAL}: Tools for Validating Plans and Planning Models},
  year         = {2026},
  howpublished = {Software repository},
  note         = {Pinned in this study at commit 3c7a1f330bdab0ba28a4762bb45c3f06c27fb6d4},
  url          = {https://github.com/KCL-Planning/VAL}
}

@misc{hinton2015distilling,
  author        = {Geoffrey Hinton and Oriol Vinyals and Jeff Dean},
  title         = {Distilling the Knowledge in a Neural Network},
  year          = {2015},
  eprint        = {1503.02531},
  archiveprefix = {arXiv},
  primaryclass  = {stat.ML},
  url           = {https://arxiv.org/abs/1503.02531}
}

@misc{salas2026correct,
  author        = {Jesus Salas},
  title         = {Correct Is Not Governed: Provenance Integrity in Agentic Workflows},
  year          = {2026},
  eprint        = {2608.12761},
  archiveprefix = {arXiv},
  primaryclass  = {cs.AI},
  url           = {https://arxiv.org/abs/2608.12761}
}
